\documentclass[10pt,twocolumn,letterpaper]{article}

\usepackage[pagenumbers]{cvpr}
\usepackage{amsmath,amssymb,mathtools}
\usepackage{graphicx}
\usepackage{booktabs,multirow,tabularx}
\usepackage{float}
\usepackage{xcolor,colortbl}
\usepackage{microtype}
\usepackage{xspace}
\usepackage{enumitem}
\usepackage{tikz}
\usetikzlibrary{arrows.meta,positioning,calc,fit}
\usepackage{url}

\definecolor{amnavy}{HTML}{24324B}
\definecolor{amviolet}{HTML}{7664D8}
\definecolor{ammagenta}{HTML}{C95B98}
\definecolor{amorange}{HTML}{E58A36}
\definecolor{amcyan}{HTML}{77BFC7}
\definecolor{amwash}{HTML}{F3F0FC}
\definecolor{amgray}{HTML}{667085}
\definecolor{hintred}{HTML}{F05A55}
\definecolor{hintgreen}{HTML}{33B978}
\definecolor{preprintink}{HTML}{B35A24}
\definecolor{preprintwash}{HTML}{FBECDC}

\newcommand{\method}{\mbox{AgentMem}\xspace}

\newcolumntype{Y}{>{\raggedright\arraybackslash}X}
\newcommand{\compactparagraph}[2][6pt]{\par\addvspace{#1}\noindent\textbf{#2}\enspace\ignorespaces}

\definecolor{cvprblue}{rgb}{0.21,0.49,0.74}
\usepackage[pagebackref,breaklinks,colorlinks,allcolors=cvprblue]{hyperref}
\def\confName{CVPR}
\def\confYear{2027}
\hypersetup{pdftitle={AgentMem: Agent-Side Memory with Steerable Action Models in Long-Horizon Manipulation},pdfauthor={Yutong Hu, Xiangming Zhu, Fengjiao Chen, Xuezhi Cao, Renaud Detry}}

\title{AgentMem: Agent-Side Memory with Steerable Action Models\\in Long-Horizon Manipulation}

\author{
\normalsize Yutong Hu$^{1,2,3}$\quad
Xiangming Zhu$^{4}$\quad
Fengjiao Chen$^{2}$\quad
Xuezhi Cao$^{2}$\quad
Renaud Detry$^{1,3,5}$ \\[1ex]
\small $^{1}$KU Leuven, Dept. Mechanical Engineering, Research unit Robotics, Automation and Mechatronics\\
\small $^{2}$Meituan Inc.\quad $^{3}$Flanders Make@KU Leuven\quad $^{4}$Shanghai Jiao Tong University\\
\small $^{5}$KU Leuven, Dept. Electrical Engineering, Research unit Processing Speech and Images
}

\begin{document}
\maketitle
\raggedbottom

\begin{abstract}
In long-horizon robot manipulation, memory is more than a record of the past: it is an evolving basis for present action and future decisions. This raises two related questions: what information should be carried forward for future decisions, and how should its implications be communicated for current action? We study both with \method, a recurrent system connecting a memory-bearing multimodal agent to a steerable action model. The agent maintains selected visual keyframes, issued subgoals, and compact facts, while conveying its current decision through subgoal language and optional image-space targets. The action model executes these commands from current RGB observations and robot state, without receiving episode history. This explicit interface allows us to vary memory inputs and command interfaces separately. Controlled ablations show that semantic records recover much of the benefit of persistent memory, while historical visual evidence provides complementary gains. Memory and steering reinforce each other: steering yields substantially larger gains with persistent history, and memory provides greater benefits through the steerable interface. Under a shared training recipe, \method achieves consistently strong performance across 42 tasks in four simulation benchmarks (RoboMME, RMBench, LIBERO-Mem, and RoboDojo-Mem), with further evaluation on three real-robot task families. Code and videos: \href{https://agent-mem-action-model.github.io/}{agent-mem-action-model.github.io}.
\end{abstract}

\begin{figure}[t]
  \centering
  \includegraphics[width=\columnwidth]{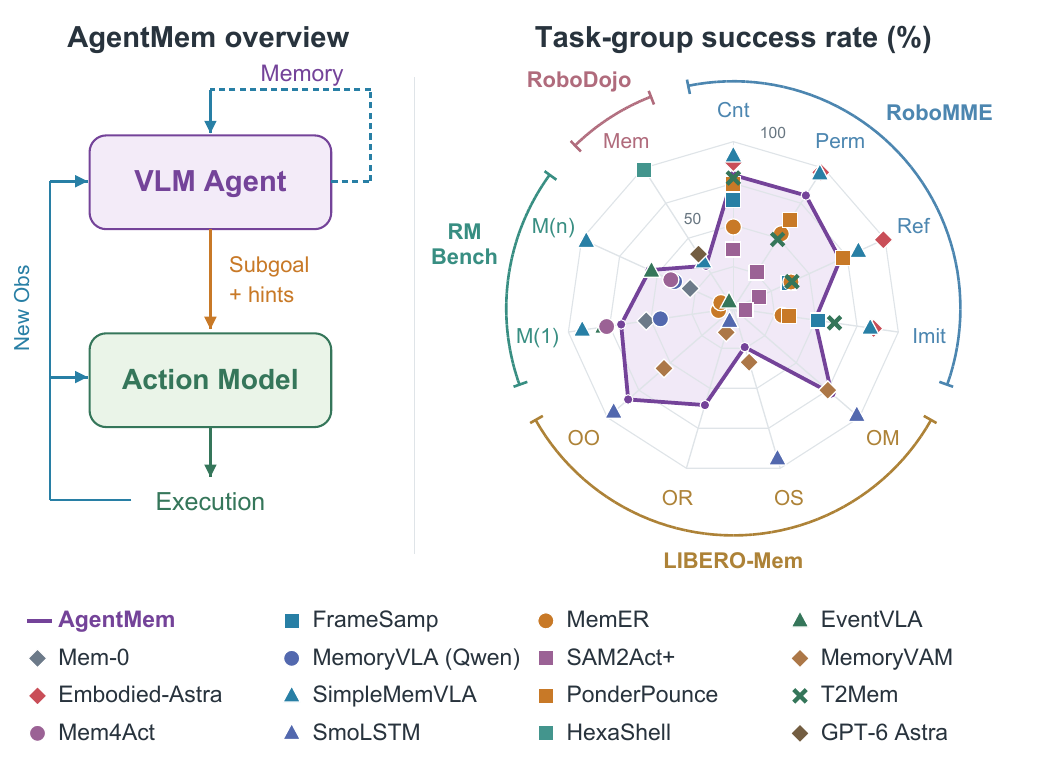}
  \caption{\textbf{Left: Overview of \method.} We share the System 1-2 hierarchy of MemER~\citep{sridhar2025memer} and MEM~\citep{torne2026mem}. Task memory resides entirely in the high-level Agent, which conveys physical intentions to the Action Model through subgoals and point-based steering. The Agent's updated memory feeds its next reasoning step. Execution supplies new observations to both models.
  \textbf{Right: Benchmark performance.} \method achieves competitive success rates across four memory benchmarks, improving over established benchmark and hierarchical baselines~\citep{robomme2026,sridhar2025memer,yang2026eventvla,chen2026rmbench,shi2025memoryvla,fang2025sam2act,jiang2026memoryvam}. Figure~\ref{fig:main-results} expands this comparison; Appendix~\ref{app:external-comparisons} provides detailed results and sources.}
  \label{fig:mini-overview}
\end{figure}

\section{Introduction}
\label{sec:intro}

Generalist robot policies have made substantial progress in following language instructions and executing manipulation skills \citep{kim2024openvla,black2025pi05}. In long-horizon manipulation, however, the correct action can depend on events no longer visible in the current scene. Consider a robot that pours parts into visually identical totes, then watches them being restacked. To retrieve the correct tote later, it must remember its contents and track its changing position. This raises two related questions: what to retain for future decisions, and how to use it to guide current action.

Existing systems connect history and control in different ways. Memory-augmented policies retain visual features or recurrent state within the action-generating model \citep{shi2025memoryvla,yang2026eventvla}. Hierarchical systems instead let a high-level model interpret history and issue instructions. MemER~\citep{sridhar2025memer} selects historical keyframes to generate language instructions for a controller operating on current observations, while MEM~\citep{torne2026mem} recurrently updates high-level language memory. SimpleMemVLA~\citep{yin2026simplememvla} conveys historical context through the embeddings and contextual hidden states of a generated subtask. Recent GPT-6 Astra~\citep{zhang2026unexpected,galbot2026astra} evaluations provide further evidence for agentic robot policies that reason over feedback and guide or generate actions. These developments motivate a question beyond where memory resides: how does the benefit of persistent memory depend on the interface through which it guides action?

We study this question through \method, a recurrent agent--action system coupling memory updates with grounded action guidance. \textbf{Across decisions}, the agent uses new observations to update selected keyframes, subgoal history, and compact facts, passing task context to its next reasoning step. \textbf{Across models}, it communicates the current decision through subgoal language and optional image-space targets, without passing stored history or history-conditioned hidden states. The action model follows these commands from current RGB observations and robot state.

Memory updates preserve context for \textbf{future} reasoning, while grounded commands specify its implications for \textbf{current} action. Subgoal language can specify a behavior while leaving its object or destination implicit. Following Steerable Policies~\citep{chen2026steerable}, we make these arguments explicit: in the tote example, the agent points to the tote identified from history. We test whether the value of this guidance depends on access to that history.

Our contributions are twofold:
\begin{itemize}[leftmargin=*,topsep=3pt,itemsep=2pt,parsep=0pt]
    \item \textbf{The interplay between memory and action.} We study what task information should be carried forward to future decisions and how it should guide current action. Controlled ablations show that persistent memory and steering reinforce each other's benefits, while historical visual evidence complements semantic records.
    \item \textbf{A strong System 1--System 2 baseline.} We present \method, combining supervised recurrent agent memory with a steerable action model through subgoal language and spatial guidance. Under a shared training recipe, it achieves consistently strong performance across four memory benchmarks~\citep{robomme2026,chen2026rmbench,liberomem2025,robodojo2026} compared with established baselines~\citep{sridhar2025memer,shi2025memoryvla,yang2026eventvla,fang2025sam2act,jiang2026memoryvam,hanyu2025slotvla,black2024pi0,black2025pi05,zhao2023act,chi2023diffusion}.
\end{itemize}

\section{Related Work}
\label{sec:related}

\compactparagraph[9pt]{Memory-augmented policies.}
Memory-augmented policies condition control on retrieved, recurrent, or compressed history~\citep{shi2025memoryvla,shah2026halo,yang2026eventvla,li2026remembervla,wang2026nativemem,qu2026lamemvla,liu2026t2mem,zhang2026mem4act,habekost2026smolstm,sun2026ltm,cherepanov2026muvla,pala2026membodied,liu2026streampi,memoact2026,prism2026,dram2026,dividremember2026,dashora2026workspace,chen2026aura,zhu2026weavela,zeng2026kemo,li2026trmvla}, spanning object-centric and spatial representations~\citep{liberomem2025,fang2025sam2act,hanyu2025slotvla,liu2026ecomem,osterberg2026unimem,zhao2026atlasvla,li2026bridgevlapp,zhao2026himem,han2026stvla,li2026soma} and execution dynamics~\citep{yuan2026motionvla,li2026fmvla,yang2026temporalflow,wang2026camp,zhou2026actioneffect,li2026trace,ding2026temporalforcing,feng2026tempo,huang2026chainvla,evomem2026,actionhistory2026}. Memory also supports joint prediction and control~\citep{jiang2026memoryvam,shi2026memoryvlapp,memorywam2026,sun2026himemwam,wang2026dimwam,li2026unimpa,zhao2026memoryplans} and cross-task skill transfer~\citep{si2026vlapro,wang2026skillmemo}. SimpleMemVLA~\citep{yin2026simplememvla} passes history-contextualized subtask representations to its action expert; PonderPounce~\citep{choi2026ponderpounce} asynchronously supplies episode context. \method instead restricts historical influence on the controller to an explicit current command, enabling separate comparisons of memory inputs and command interfaces.

\compactparagraph[9pt]{Agent-side memory.}
Agent-side memory and stateful harnesses maintain task context for planning~\citep{lei2026robomemory,ji2026hime,ledger2026,xu2026explicitmemory,zhao2026hybridmemory,banwasi2026messymem,pal2026,li2026roboharness,vo2026codegraphvlp}. MemER~\citep{sridhar2025memer} learns keyframe selection and issues language subtasks to a current-observation controller; its modality ablation finds no benefit from predicted subtask text on its evaluated tasks. MEM~\citep{torne2026mem} uses recurrent language summaries and combines them with low-level video memory. Its compression ablations and memory-baseline comparisons use custom real-robot tasks rather than shared memory benchmarks. Mem-0~\citep{chen2026rmbench} likewise maintains history in planning and execution. \method maintains visual and semantic history through a supervised Agent-side memory loop, without passing history to the Action Model. Across four public memory benchmarks and controlled cross-interface ablations, we study what is retained and how its benefit depends on guidance to the controller.

\compactparagraph[9pt]{Communication for action.}
Hierarchical manipulation connects planning and execution through language, code, or spatial commands~\citep{ahn2022saycan,liang2023codeaspolicies,huang2023voxposer,huang2024rekep,belkhale2024rth,shi2025hirobot,li2025hamster,zheng2025tracevla,liu2026lohomanip,feng2025reflective,li2025have,chen2026volo,yang2025agenticrobot,long2026vista,li2026roboclaw,liu2026goal2skill,lei2026sparkvla,xu2026bcp,cai2026tauvla}. EPoG~\citep{yang2026epog} combines graph-based global planning with LLM-based situated replanning to integrate exploration and sequential manipulation under partial observability. Agentic harnesses further adapt this coordination using execution feedback~\citep{harnessvla2026,chen2026openeta,yang2026incontextvla,tang2026adahvla,xu2026baton,sera2026}. Steerable Policies~\citep{chen2026steerable} and FineVLA~\citep{hu2026finevla} improve fine-grained command following. GPT-6 Astra evaluations~\citep{zhang2026unexpected,galbot2026astra} and Astranomy~\citep{yang2026astranomy} explore direct and agentic control with external models. 2AM~\citep{hu2026twoam} preliminarily validated the steering contract on LIBERO-Mem, without specifying a learned Agent-side memory-update mechanism or evaluating memory--steering interactions. \method develops supervised recurrent Agent memory and studies its interaction with action guidance through retrained controls, including target-specific language, across benchmarks.

\section{Method}
\label{sec:method}

\begin{figure*}[!t]
\centering
\includegraphics[width=\textwidth]{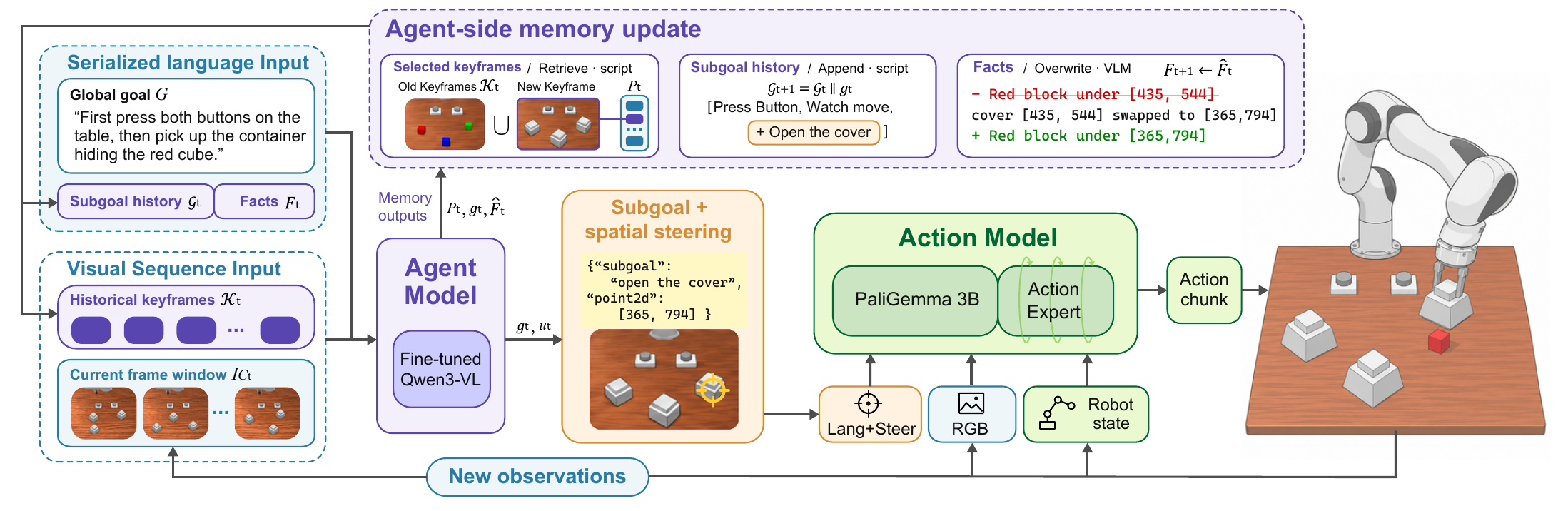}
\caption{\textbf{Agent-side memory over a steerable Action Model.} The high-level Agent reads the goal $G$, subgoal history $\mathcal G_t$, and facts $F_t$ as serialized language, together with the visual sequence combining the current frame window $I_{C_t}$ and historical keyframes $\mathcal K_t$. It outputs a current subgoal $g_t$, point-based steering $u_t$, selected frame indices $P_t$, and rewritten facts $\widehat F_t$. Memory recurs through the Agent's own outputs: external scripts retrieve and retain the selected frames and append the issued subgoal, while the facts are directly rewritten. These updated memories return as inputs at the next decision. The Action Model receives only the current subgoal with steering $(g_t,u_t)$, current RGB, and robot state, not the stored history. It executes an action chunk, producing fresh observations for both models and closing the loop.}
\label{fig:teaser}
\end{figure*}

\method turns the implications of task history into commands for current action. The Agent maintains and interprets memory, while the Action Model executes from current observations and a grounded command. Their interface combines subgoal language with optional image-space steering targets. Agent outputs update the memory used in subsequent decisions, and executed actions produce fresh observations, forming a recurrent closed-loop system.

\subsection{Closed-loop Agent--Action system}

Let $G$ denote the overall goal, $o_t=(\mathbf I_t,q_t)$ the current robot observation, and $A_t$ an action chunk. Here $\mathbf I_t$ contains the Action Model's current camera images and $q_t$ is the robot state. We index Agent decision rounds by $t$; $A_t$ contains the control actions predicted for that round, not one action per Agent call. A current-observation policy models $\pi(A_t\mid o_t,G)$. For memory-dependent tasks, however, the current scene may not determine the correct object, destination, or stage, motivating conditioning on the observation history $o_{0:t}$.

Let $g_t$ denote the current subgoal language and $u_t$ the optional image-space steering targets. Together, $(g_t,u_t)$ form the current command. Directly conditioning the joint policy $\pi(A_t,g_t,u_t\mid o_{0:t},G)$ on an ever-growing visual history becomes increasingly costly in memory and inference time. MemER~\citep{sridhar2025memer} and MEM~\citep{torne2026mem} identify this scaling problem and use selected visual evidence or compressed semantic records to retain useful context. Other approaches maintain learned memory within the action-generating model \citep{shi2025memoryvla}. We instead retain task history on the Agent side and expose only its current grounded decision to the Action Model.

The Agent accesses visual history through a \emph{current frame window} and \emph{historical keyframes}. Let $I_i$ be a global-view RGB frame with episode index $i$. The current window $I_{C_t}$ contains frames indexed by $C_t$, while $\mathcal K_t$ contains historical keyframe images. Their union forms the Agent's visual input, $\mathcal I_t=I_{C_t}\cup\mathcal K_t$.
The combined images are presented in temporal order, with repeated frame indices included only once. The Agent also reads subgoal history $\mathcal G_t$ and compact facts $F_t$, when enabled.

\begin{samepage}
With these inputs, we factorize the policy as
\begin{equation}
\begin{aligned}
&\pi(A_t,g_t,u_t\mid o_{0:t},G)\\
&\quad\approx
\underbrace{\pi_{\mathrm{Action}}(A_t\mid o_t,g_t,u_t)}_{\text{execute the current decision}}\\
&\qquad\times
\underbrace{\pi_{\mathrm{Agent}}(g_t,u_t\mid G,\mathcal I_t,\mathcal G_t,F_t)}_{\text{reason over task history}}.
\end{aligned}
\label{eq:decomposition}
\end{equation}
\end{samepage}
The decomposition separates responsibilities, not the influence of history: actions remain history-informed through $(g_t,u_t)$, although the Action Model does not receive the stored history. New observations allow the Agent to reconsider the command before subsequent execution.

\subsection{Recurrent Agent-side memory}

As shown in Figure~\ref{fig:teaser}, Agent-side memory is maintained through recurrent interaction: each call reads records produced by earlier calls and predicts both a current command and memory content for the next call:
\begin{equation}
\begin{aligned}
&(g_t,u_t,P_t,\widehat F_t)\\
&\quad\sim\pi_{\mathrm{Agent}}
(\,\cdot\mid G,\mathcal I_t,\mathcal G_t,F_t).
\end{aligned}
\label{eq:agent}
\end{equation}
The selected indices $P_t\subseteq C_t$ identify frames worth retaining from the current window. The VLM also generates an updated facts collection $\widehat F_t$ for tasks that use this path. Its outputs therefore express both a current physical intention and the information to carry into subsequent decisions.

\compactparagraph{Historical keyframes.}
Following the use of selected visual evidence in MemER~\citep{sridhar2025memer}, keyframes preserve observations that may disappear from the current scene, such as an object's color before occlusion or a reference demonstration. Selected indices retrieve images from the current window, giving the logical update $\mathcal K_{t+1}=\mathcal K_t\cup\operatorname{Retrieve}(I_{C_t},P_t)$. Only selected window frames enter persistent memory; other frames need not survive the window. Appendix~\ref{app:visual-runtime} details storage and redundancy handling.

\compactparagraph{Subgoal history.}
The list $\mathcal G_t$ records previously issued subgoals. It gives the Agent a history of its decisions while new observations reveal their outcomes. An issued instruction does not establish successful execution: the Agent may continue, retry, or advance after observing the result.

\compactparagraph{Compact facts.}
Some tasks require a small amount of information to persist across stages, such as initial colors, a reference identity, or an observed ordering. As in recurrent language memory \citep{torne2026mem}, the VLM reads $F_t$ and generates a revised collection $\widehat F_t$. Unlike the subgoal history, this record is rewritten rather than continually appended. Facts keep relevant information explicit, while keyframes preserve evidence the Agent can inspect again.

Together, these outputs construct the three memory-bearing inputs for the next round:
\begin{equation}
\begin{alignedat}{2}
\mathcal I_{t+1}&=I_{C_{t+1}}\cup\mathcal K_{t+1}
&\quad&\text{script-maintained}\\
\mathcal G_{t+1}&=\mathcal G_t\Vert g_t
&\quad&\text{script-maintained}\\
F_{t+1}&\leftarrow\widehat F_t
&\quad&\text{VLM-rewritten}
\end{alignedat}
\label{eq:memory-update}
\end{equation}
External scripts assemble the visual input and append the issued subgoal, with $\Vert$ denoting list append. For tasks using facts, the VLM generates $\widehat F_t$ to replace $F_t$; the runtime stores the replacement. These records return with a new current frame window at the next round, closing the memory feedback path. Recurrence resides in externally maintained records and repeated Agent calls, not an additional recurrent neural module. Appendix~\ref{app:semantic-runtime} specifies the bookkeeping.

\subsection{Steerable Action Model}
\label{sec:contract}

Building on spatially steerable policies \citep{chen2026steerable}, subgoal language $g_t$ specifies a local behavior, while image-space steering targets $u_t$ provide optional labeled 2D grasp, place, or move targets. After tracking the tote containing the requested parts through restacking, for example, the Agent can request ``Pick up the tote'' and point to it in the current image. The memory remains with the Agent; the command conveys the target needed for this action.

Steering targets refer to the current global RGB view; unneeded targets are omitted. They guide learned motion rather than prescribe a trajectory. As specified in Eq.~\ref{eq:decomposition}, the Action Model conditions only on $o_t$, $g_t$, and $u_t$. It receives neither historical keyframes, the prior subgoal list, nor facts, and carries no persistent episode history between calls.

Execution closes the physical feedback path: observations produced during an action chunk enter the next current frame window, allowing the Agent to continue, revise, or replace its command based on observed progress.

\section{Implementation}
\label{sec:implementation}

\subsection{Models and observation interface}

The Agent is a fine-tuned Qwen3-VL~\citep{bai2025qwen3vl} 4B model. Its input begins with the global-view RGB sequence $\mathcal I_t$, resized to $224\times224$ pixels, followed by serialized JSON records of previous subgoals and facts, and the overall goal $G$. The images combine the current frame window with historical keyframes as defined in Section~\ref{sec:method}. The Agent generates a JSON response containing the current subgoal $g_t$, image-space steering targets $u_t$, keyframe selections $P_t$, and updated facts $\widehat F_t$ when applicable.

The Action Model uses a 3B PaliGemma backbone and a 300M action expert, fine-tuned from \texttt{Pi-base} following $\pi_0$~\citep{black2024pi0}. It receives all available camera views at a single timestep, robot state $q_t$, and $L_t=\operatorname{Serialize}(g_t,u_t)$: subgoal language and image-space steering targets serialized into one structured language instruction. Coordinates are normalized to $[0,1000]^2$ in the global view. Episode history remains on the Agent side. Model settings and the serialization contract are detailed in Appendices~\ref{app:models} and~\ref{app:interface}.

\subsection{Supervised training}

Demonstrations supervise memory updates and grounded command generation in the Agent, and command-conditioned execution in the Action Model. The two models are trained separately, using each simulation benchmark's provided per-task demonstration counts. Across the four simulation benchmarks, the shared recipe trains the Agent for three epochs and the Action Model for ten.

\begin{figure*}[t]
\centering
\includegraphics[width=\textwidth]{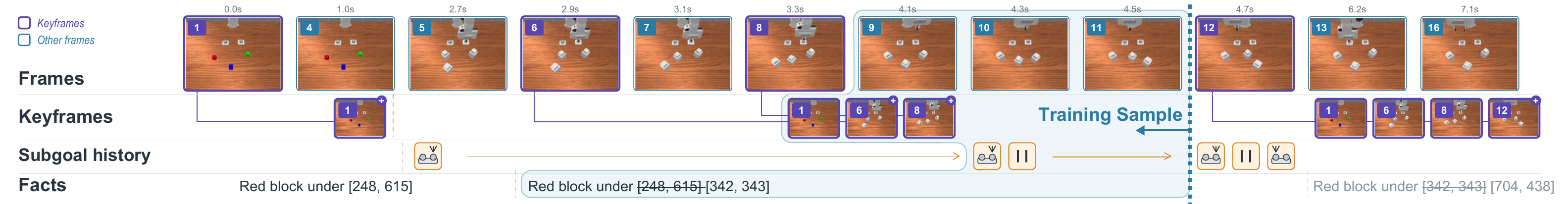}
\caption{\textbf{Causal sampling of training data.} Given an expert trajectory annotated with keyframes, subgoals, and facts, each training sample combines its current frame window with the keyframes and subgoal history retained before the sampled decision, and the latest facts valid at that time. Light blue shading marks one such sampled input. Later frames and memory updates are excluded from the inputs to match the causal information available at inference and prevent future annotations from leaking into the training context.}
\label{fig:training}
\end{figure*}

\compactparagraph{Subtasks and keyframes.}
We start from expert trajectories segmented into subtasks. For datasets without these annotations, we use a code-agent-assisted procedure to annotate subtask segments. Subtask transitions provide candidate keyframe locations, which are refined using trajectory heuristics such as gripper opening or closing and local minima of the end-effector trajectory. Subtask segments supply subgoal-language labels, while the selected frames supervise retention of visual evidence. Data sources and annotation procedures are detailed in Appendices~\ref{app:data-splits} and~\ref{app:annotation}.

\compactparagraph{Steering targets and facts.}
Where object geometry and camera calibration are available, we project the target object's 3D centroid into the global image. Otherwise, SAM~\citep{kirillov2023sam} provides a target mask whose weighted pixel centroid defines the point. Targets are normalized by image dimensions and paired with the corresponding steering role. Geometry and SAM are used only for offline supervision; neither is available to the deployed system. Dataset-specific annotation rules are given in Appendix~\ref{app:spatial-labels}.

Facts are annotated as concise JSON records of task-relevant information, including the objects being swapped, their initial positions or colors, and the number of actions already completed. Each record reflects the observed demonstration prefix rather than future outcomes. Appendix~\ref{app:facts} provides illustrative records.

\compactparagraph{Agent fine-tuning.}
We construct training batches by sampling sliding windows from expert trajectories. Each example contains the current frame window, historical keyframes annotated for retention before the current decision, the preceding expert subgoal sequence, and facts available at the current decision. The output is serialized JSON supervising the four predictions in Eq.~\ref{eq:agent}. Let $x_t^*$ denote this multimodal input and $y_t^*$ the target token sequence. We train the Agent with next-token prediction:
\begin{equation}
\mathcal L_{\mathrm{Agent}}=
-\mathbb E_{(x_t^*,y_t^*)}
\sum_{j=1}^{|y_t^*|}
\log p_\theta(y_{t,j}^*\mid x_t^*,y_{t,<j}^*).
\label{eq:agent-sft}
\end{equation}
Figure~\ref{fig:training} makes the causal boundary explicit. For each sampled window, we reconstruct the previously retained keyframes, preceding subgoal sequence, and latest available facts, rather than reuse the final episode records. Repeated frame indices are included only once. Current keyframe-selection and facts-update labels supervise the output; they are not inserted retroactively into that sample's memory inputs. Previous expert subgoals provide supervised decision context, not online success signals. Sampling and optimization are detailed in Appendices~\ref{app:causal-sampling} and~\ref{app:agent-training}.

\compactparagraph{Action Model fine-tuning.}
We fine-tune the Action Model with the flow-matching objective of $\pi_0$~\citep{black2024pi0}, conditioned on current multiview RGB, robot state, and the structured language $L_t$ encoding $g_t$ and $u_t$. The objective and optimization details are provided in Appendix~\ref{app:action-training}.

\subsection{Memory and closed-loop execution}

At each decision round, the Agent receives a fresh current window of eight frames evenly sampled from the preceding execution, together with memory maintained from its own previous outputs. Training and deployment therefore share the same recurrent input/output structure, although training histories come from annotations and deployment histories are model-generated.

An external parser reads the generated JSON. If parsing fails, the runtime requests regeneration; only a successfully parsed response is used to update memory or issue the next command. Across all experiments, no Agent decision incurred three consecutive parsing failures. The selected window positions are mapped to frame indices and the corresponding images are retained, with temporal deduplication and the initial frame preserved. We impose no fixed memory budget; the largest keyframe set observed in our experiments contains 24 frames. The script appends the subgoal and compresses consecutive repeated descriptions, while the VLM-generated facts collection overwrites the previous facts, as in Eq.~\ref{eq:memory-update}.

The subgoal language and steering targets are serialized into $L_t$ for the Action Model, which executes one action chunk while new images enter the current frame window. The next Agent call receives this window and the updated memory. This cycle links memory-informed decisions to observed execution without passing stored history to the Action Model or simulator task state to the Agent. Appendix~\ref{app:execution} collects parsing, timing and termination details.

\section{Experiments}
\label{sec:experiments}

We first examine how memory and command interfaces affect performance through controlled ablations (Table~\ref{tab:ablation}), then assess the design across four simulation benchmarks and three real-robot task families.

\begin{table*}[t]
\centering\footnotesize
\setlength{\tabcolsep}{2pt}
\caption{\textbf{Memory and steering across six tasks.} SR (\%), mean $\pm$ sample standard deviation across three evaluation seeds, with 50 trials per task per seed. The final column averages six tasks within each seed. History excludes the current window; commands supplement current RGB and robot state. Variants are retrained; $g_t^{+}$ denotes expanded target-specific language. Bold marks the highest mean per column.}
\label{tab:ablation}
\begin{tabularx}{\textwidth}{@{}Yccrrrrrrr@{}}
\toprule
& & & \multicolumn{3}{c}{RoboMME} & \multicolumn{3}{c}{RMBench} & Six-task\\
\cmidrule(lr){4-6}\cmidrule(lr){7-9}
Question & History & Command & \shortstack{Button\\UnmaskSwap} & \shortstack{Video\\PlaceOrder} & \shortstack{Pattern\\Lock} & \shortstack{Blocks\\Ranking} & Swap T & \shortstack{Press\\Button} & mean\\
\midrule
\rowcolor{amwash}
\textbf{Full system (reference)} & $(\mathcal K_t,\mathcal G_t,F_t)$ & $(g_t,u_t)$ & $\mathbf{66.7}$\,{\scriptsize$\pm3.1$} & $\mathbf{78.0}$\,{\scriptsize$\pm4.0$} & $\mathbf{61.3}$\,{\scriptsize$\pm4.2$} & $88.0$\,{\scriptsize$\pm6.0$} & $\mathbf{78.7}$\,{\scriptsize$\pm6.1$} & $\mathbf{29.3}$\,{\scriptsize$\pm4.2$} & $\mathbf{67.00}$\,{\scriptsize$\pm1.53$}\\
Does history help? & $\varnothing$ & $(g_t,u_t)$ & $17.3$\,{\scriptsize$\pm3.1$} & $20.7$\,{\scriptsize$\pm3.1$} & $12.7$\,{\scriptsize$\pm3.1$} & $77.3$\,{\scriptsize$\pm12.1$} & $12.7$\,{\scriptsize$\pm2.3$} & $2.0$\,{\scriptsize$\pm0.0$} & $23.78$\,{\scriptsize$\pm1.84$}\\
Images beyond semantics? & $(\mathcal G_t,F_t)$ & $(g_t,u_t)$ & $53.3$\,{\scriptsize$\pm4.2$} & $76.0$\,{\scriptsize$\pm4.0$} & $51.3$\,{\scriptsize$\pm3.1$} & $84.7$\,{\scriptsize$\pm11.5$} & $60.0$\,{\scriptsize$\pm8.7$} & $16.0$\,{\scriptsize$\pm0.0$} & $56.89$\,{\scriptsize$\pm2.22$}\\
Semantics beyond images? & $\mathcal K_t$ & $(g_t,u_t)$ & $29.3$\,{\scriptsize$\pm3.1$} & $28.0$\,{\scriptsize$\pm4.0$} & $8.0$\,{\scriptsize$\pm2.0$} & $84.7$\,{\scriptsize$\pm4.2$} & $15.3$\,{\scriptsize$\pm3.1$} & $3.3$\,{\scriptsize$\pm1.2$} & $28.11$\,{\scriptsize$\pm1.26$}\\
Steering without history? & $\varnothing$ & $g_t$ & $4.0$\,{\scriptsize$\pm2.0$} & $8.0$\,{\scriptsize$\pm2.0$} & $10.7$\,{\scriptsize$\pm3.1$} & $72.7$\,{\scriptsize$\pm6.1$} & $10.0$\,{\scriptsize$\pm3.5$} & $1.3$\,{\scriptsize$\pm2.3$} & $17.78$\,{\scriptsize$\pm1.58$}\\
Steering with full history? & $(\mathcal K_t,\mathcal G_t,F_t)$ & $g_t$ & $0.7$\,{\scriptsize$\pm1.2$} & $8.7$\,{\scriptsize$\pm3.1$} & $28.7$\,{\scriptsize$\pm3.1$} & $\mathbf{90.0}$\,{\scriptsize$\pm2.0$} & $6.7$\,{\scriptsize$\pm3.1$} & $2.7$\,{\scriptsize$\pm1.2$} & $22.89$\,{\scriptsize$\pm0.77$}\\
Beyond richer language? & $(\mathcal K_t,\mathcal G_t,F_t)$ & $g_t^{+}$ & $15.3$\,{\scriptsize$\pm3.1$} & $26.7$\,{\scriptsize$\pm3.1$} & $6.0$\,{\scriptsize$\pm2.0$} & $82.7$\,{\scriptsize$\pm3.1$} & $14.0$\,{\scriptsize$\pm2.0$} & $2.7$\,{\scriptsize$\pm1.2$} & $24.56$\,{\scriptsize$\pm0.84$}\\
\bottomrule
\end{tabularx}
\end{table*}

\subsection{Benchmarks and evaluation protocol}

We evaluate RoboMME~\citep{robomme2026} (16 tasks, 50 trials per task), RMBench~\citep{chen2026rmbench} (10 tasks, 100 trials per task), LIBERO-Mem~\citep{liberomem2025} (10 tasks, 100 trials per task), and RoboDojo-Mem~\citep{robodojo2026} (6 tasks, 50 trials per task). Figure~\ref{fig:main-results} presents the main comparison; Appendix~\ref{app:fixed-results} reports per-task results.

RMBench comparisons use the common nine-task subset excluding Place-Mat; the rightmost column also reports ten-task means when available. On LIBERO-Mem, SR enforces the required sequence and counts, and completion measures subgoal progress. Its original baselines~\citep{liberomem2025} report completion, not SR. Appendix~\ref{app:evaluation} provides evaluator settings, metric definitions and trial accounting.

\subsection{Memory and interface ablations}
\label{sec:ablations}

Subgoal language may leave task arguments implicit; steering provides a way to specify them more precisely \citep{chen2026steerable}. We test how its benefit depends on task history. Table~\ref{tab:ablation} spans trial-and-error ordering (Blocks Ranking), identity tracking under rearrangement (the swap tasks), extended progress tracking (Press Button), and sequence dependencies (PatternLock and VideoPlaceOrder). History is full $(\mathcal K_t,\mathcal G_t,F_t)$, semantic-only $(\mathcal G_t,F_t)$, visual-only $\mathcal K_t$, or absent ($\varnothing$); all retain the current window. Commands are $(g_t,u_t)$, $g_t$, or expanded language $g_t^{+}$, always alongside current RGB and robot state.

Agent memory variants and Action Model interface variants are retrained. Within each benchmark, the full-memory steering and original language-only conditions share an Agent checkpoint. We report mean and sample standard deviation across three evaluation seeds, with 50 trials per task per seed in both benchmarks. Six-task means weight tasks equally within each seed. These subset results are separate from full-suite scores.

\compactparagraph{Persistent history supports current decisions.}
Across six tasks, removing persistent history reduces mean SR from 67.00\% to 23.78\% with steering. The current window alone therefore cannot recover full-system performance.

\compactparagraph{Memory and steering provide complementary gains.}
With full memory, steering adds 44.1 points over the retrained language-only controller; without persistent history, it adds 6.0. Conversely, memory adds 43.2 points through steering versus 5.1 through language alone. The steering interaction holds on both RoboMME (56.0 versus 9.3 points) and RMBench (32.2 versus 2.7). Neither a richer interface nor persistent history alone recovers the full system's performance on these tasks.

The gains are task-dependent: steering offers no mean improvement on Blocks Ranking (88.0\% versus 90.0\% with full-memory language). Language already achieves high success, leaving limited headroom, and block colors allow explicit target references, reducing the need for spatial disambiguation. In contrast, steering adds 72.0 points on Swap T and 26.7 on Press Button. Such variation is compatible with full memory and steering achieving the highest six-task mean on all three evaluation seeds.

\compactparagraph{Visual evidence complements semantic records.}
Semantic records alone reach 56.89\% mean SR, versus 28.11\% for historical images alone and 23.78\% for the current window. Adding images to semantic records provides another 10.1 points; adding semantic records to images provides 38.9. Thus, semantic records recover much of the memory benefit, while visual evidence supplies complementary gains overall. Images alone help on Blocks Ranking (84.7\% versus 77.3\% with the current window), but not PatternLock (8.0\% versus 12.7\%).

\compactparagraph{Spatial guidance or richer descriptions?}
Steering adds target information as well as a spatial interface. To test whether richer descriptions alone recover its benefit, we introduce expanded subgoal descriptions $g_t^{+}$. An external code agent uses ground-truth target information and current images to relabel dataset subgoals so that each description uniquely identifies its intended target. We retrain both the full-memory Agent and the language-only Action Model on these descriptions; annotation details are in Appendix~\ref{app:ablations}. Mean SR reaches 24.56\%, versus 22.89\% with original language and 67.00\% with steering. Richer language therefore does not recover spatial guidance's benefits under this training setup. The language-only Action Model may struggle to ground the expanded descriptions; Appendix~\ref{app:language-grounding} discusses this possibility. This is consistent with Steerable Policies~\citep{chen2026steerable}, which finds that spatially grounded commands can improve target selection when language-only subtask commands fail, helping translate high-level intent into reliable execution.

See Appendix~\ref{app:ablations} for training and evaluation details.

\subsection{A strong cross-benchmark baseline}

\begin{figure*}[t]
\centering
\includegraphics[width=\textwidth]{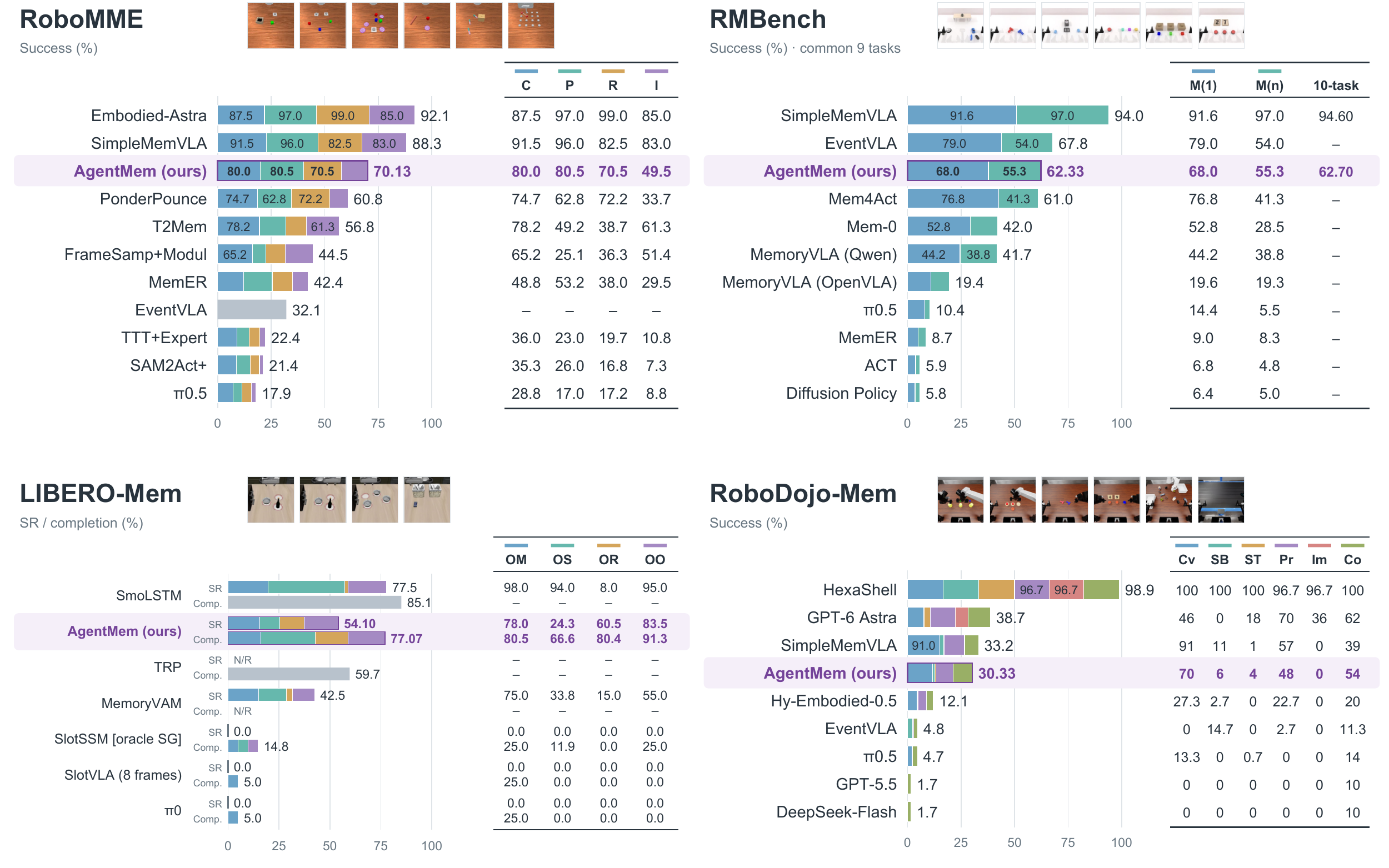}
\caption{\textbf{Results across four memory benchmarks.}
Purple highlights \method. Bars sum task-weighted contributions; adjacent tables give raw group rates. C/P/R/I denote Counting/Permanence/Reference/Imitation in RoboMME~\citep{robomme2026}. RMBench~\citep{chen2026rmbench} bars use the common nine tasks; the rightmost column additionally reports ten-task means when available. LIBERO-Mem~\citep{liberomem2025} pairs SR (success rate, upper) and completion (lower) for each method; OM/OS/OR/OO label its four disjoint task families. RoboDojo~\citep{robodojo2026} columns are Cover, Swap blocks, Swap T, Press, Imitate, and Conveyor pick. Gray indicates that a subgroup breakdown is unavailable in the source report; N/R denotes an unreported metric. Appendix~\ref{app:external-comparisons} provides full per-task comparisons and source protocols.}
\label{fig:main-results}
\end{figure*}

Across all four simulation benchmarks, we train separate models with one recipe: three VLM Agent epochs and ten VLA Action Model epochs. Other methods use the best reported configurations from their original papers or benchmark reproductions, not our retrainings. Figure~\ref{fig:main-results} compares \method with benchmark-native baselines and recent methods, including preprints available by November 2026. Appendix~\ref{app:external-comparisons} provides full per-task comparisons and protocol details.

\compactparagraph{Gains over benchmark and hierarchical baselines.}
\method exceeds the deployable baselines in each benchmark's original report. It reaches 70.13\% SR versus 44.51\% for FrameSamp+Modul~\citep{robomme2026} on RoboMME, 62.33\% versus 42.0\% for Mem-0~\citep{chen2026rmbench} on RMBench, and 30.33\% versus 12.11\% for Hy-Embodied-0.5~\citep{robodojo2026} on RoboDojo Memory. LIBERO-Mem completion reaches 77.07\%, exceeding even SlotSSM~\citep{liberomem2025} with oracle subgoal embeddings (14.8\%). Against the strong System 1--System 2 baseline MemER~\citep{sridhar2025memer}, \method improves SR from 42.38\% to 70.13\% on RoboMME and from 8.7\% to 62.33\% on RMBench. On RMBench, \method trails EventVLA~\citep{yang2026eventvla} (62.33\% versus 67.8\%), but substantially outperforms it on the other two shared benchmarks: RoboMME (70.13\% versus 32.1\%) and RoboDojo-Mem (30.33\% versus 4.78\%). \method also exceeds PonderPounce~\citep{choi2026ponderpounce} (60.83\%) and T$^2$Mem~\citep{liu2026t2mem} (56.83\%) on RoboMME, Mem4Act~\citep{zhang2026mem4act} (61.0\%) on RMBench, and TRP~\citep{aktas2026trp} (59.7\% completion) on LIBERO-Mem.

\compactparagraph{Where the gains arise.}
\method preserves the visual-recall strengths demonstrated by the keyframe-based hierarchy in MemER~\citep{sridhar2025memer}: ButtonUnmask improves from 72\% to 94\%, and PickHighlight from 70.67\% to 78\%. Gains extend to changing spatial bindings: ButtonUnmaskSwap rises from 21.33\% to 64\%, VideoUnmaskSwap from 38\% to 64\%, and RMBench Swap T from 7\% to 76\%. Consistent with this pattern, the separate three-seed Swap T ablation reaches 60.0\% with semantic records versus 12.7\% with the current window and 78.7\% with full memory. On RoboMME, gains over FrameSamp+Modul span Counting, Permanence and Reference, while Imitation remains slightly lower (49.50\% versus 51.39\%). LIBERO-Mem improves completion over its original baselines across all four task families; tasks requiring the robot to remember which basket it filled reach 69\% and 98\% SR. RoboDojo gains concentrate on Cover, Conveyor and Press (70\%, 54\%, 48\%), while swapping and imitation remain challenging.

Several methods report higher per-benchmark results: Embodied-Astra~\citep{jedlicki2026embodiedastra} on RoboMME (92.13\%), SimpleMemVLA~\citep{yin2026simplememvla} on RMBench (94.0\%), SmoLSTM~\citep{habekost2026smolstm} on LIBERO-Mem (77.5\% SR; 85.1\% completion), and HexaShell~\citep{gao2026hexashell} on RoboDojo Memory (98.89\%). These works include multi-benchmark evaluations, but none reports results across all four memory suites studied here. Their available results provide per-benchmark context; \method's contribution is a consistently strong baseline across these suites under one training recipe. Appendix~\ref{app:comparison-sources} tabulates the reported results collected in our comparison.

\compactparagraph{Runtime efficiency.}
Using the same RoboMME setup and high-level VLM and low-level VLA backbones as \method, our MemER~\citep{sridhar2025memer} implementation takes 0.707\,s per high-level call on an A100 80GB; its original report gives 0.787\,s on an RTX 4090. Adding semantic records introduces 128 input tokens on average, or 14.2\% of the visual-token budget, increasing latency to 1.016\,s. With caching and compiled inference, \method achieves 0.812\,s on one A100 and 0.663\,s on two, retaining sub-second responses with richer memory. Agent solutions using external GPT-6 Astra report substantially longer call latencies: Embodied-Astra~\citep{jedlicki2026embodiedastra} reports 18.3\,s on RoboMME, while Astranomy~\citep{yang2026astranomy} reports 18.56\,s for its direct policy on RoboLab-120.

\subsection{Real-robot evaluation}
\label{sec:real}

\begin{figure}[!t]
\centering
\includegraphics[width=\linewidth]{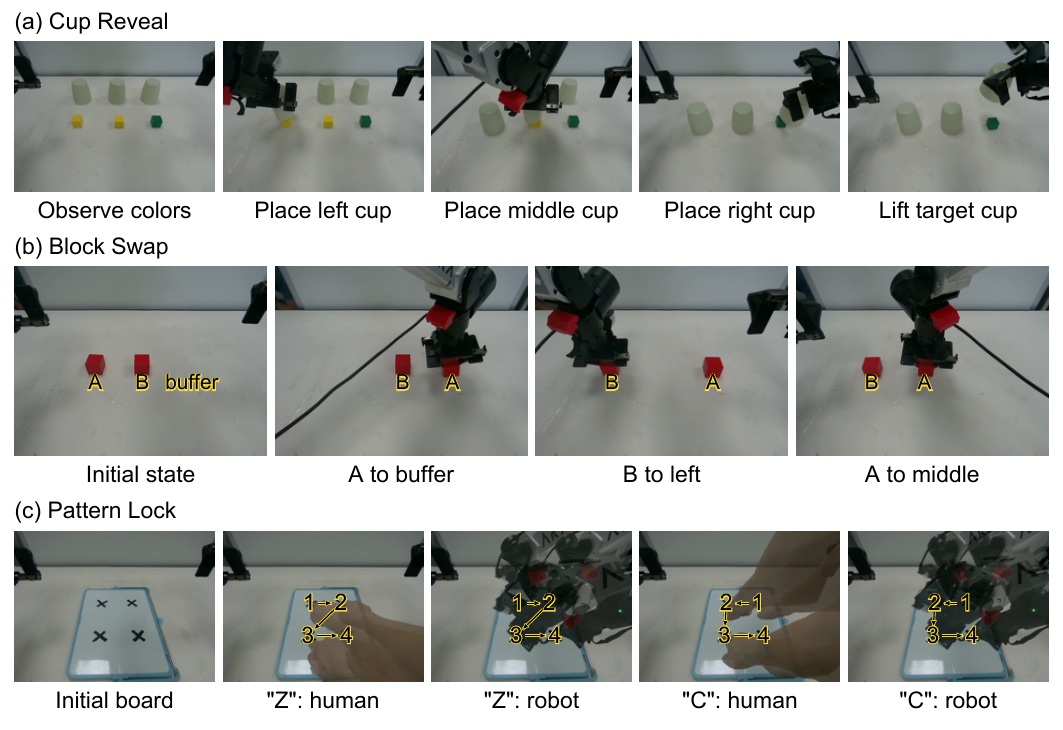}
\caption{\textbf{Real-robot tasks.} (a) Cup Reveal: remember the observed block colors, then lift the cup covering the green block. (b) Block Swap: exchange two visually similar blocks using a buffer location. (c) Pattern Lock: reproduce a demonstrated unlock sequence by touching four marked points in order, for example in a C or Z pattern.}
\label{fig:real-robot-tasks}
\vspace{8pt}
\begingroup
\centering
\small
\setlength{\tabcolsep}{2.2pt}
\captionof{table}{\textbf{Real-robot results.} SR: successful/total trials; Comp.: subgoal completion. Best values, including ties, are bold.}
\label{tab:real}
\begin{tabularx}{\linewidth}{@{}Xrrrrrr@{}}
\toprule
& \multicolumn{2}{c}{\method} & \multicolumn{2}{c}{$\pi_{0.5}$} & \multicolumn{2}{c}{$\pi_{0.5}$ (8 frames)} \\
\cmidrule(lr){2-3}\cmidrule(lr){4-5}\cmidrule(l){6-7}
Task & SR & Comp. & SR & Comp. & SR & Comp. \\
\midrule
Cup Reveal & \cellcolor{amwash}\textbf{9/15} & \cellcolor{amwash}\textbf{90.0\%} & 5/15 & 83.3\% & 5/15 & 83.3\% \\
Block Swap & \cellcolor{amwash}\textbf{15/15} & \cellcolor{amwash}\textbf{100\%} & 0/15 & 44.4\% & \textbf{15/15} & \textbf{100\%} \\
Pattern Lock & \cellcolor{amwash}\textbf{15/16} & \cellcolor{amwash}\textbf{93.8\%} & 0/16 & 7.8\% & 0/16 & 15.6\% \\
\bottomrule
\end{tabularx}
\endgroup
\end{figure}

We use EVA-Client~\citep{yang2026evaclient} as the deployment and evaluation platform for our real-robot experiments.

We evaluate three real-robot tasks (Figure~\ref{fig:real-robot-tasks}): \textbf{Cup Reveal}, remembering block colors before occlusion to uncover the specified target; \textbf{Block Swap}, exchanging two visually similar blocks through a buffer while tracking their identities; and \textbf{Pattern Lock}, reproducing a human-demonstrated sequence through four marked locations. The illustrated Z and C patterns share the same locations but differ in visitation order. We conduct 15 trials each for Cup Reveal and Block Swap, and one trial for each of 16 non-crossing patterns. We compare against single-frame and eight-frame $\pi_{0.5}$~\citep{black2025pi05}; the latter uses the eight latest visual frames during both training and inference. All tested Pattern Lock sequences appear in training, but a human taps out the target sequence anew for each evaluation episode, providing online visual evidence to remember. See Appendix~\ref{app:real-robot} for configurations and scoring.

All three tasks are designed to be non-Markovian under single-frame observations: the correct target or next step depends on earlier events. In Cup Reveal, $\pi_{0.5}$ tends to lift the rightmost cup, consistent with its 5/15 (33.3\%) success across the three configurations; eight-frame input leaves success unchanged, versus 9/15 for \method (Table~\ref{tab:real}). In Block Swap, seven historical frames plus the current observation contain the identity and stage information needed for the next action, making the task effectively Markovian with respect to this window in our setup. Success consequently rises from 0/15 to 15/15, matching \method. The advantage of \method remains on the longer, sequence-dependent Pattern Lock task, where it completes 15/16 patterns versus 0/16 for both baselines. Persistent memory thus helps when task-relevant evidence extends beyond the available window, while a short window suffices for Block Swap. Supplementary videos show task performance together with the Agent's reasoning and the Action Model's execution.

\section{Discussion and Conclusion}
\label{sec:discussion}

\method combines recurrent Agent-side memory with grounded action guidance. The Agent maintains task context and guides an Action Model without episode history through language and spatial targets. Evaluation across four simulation benchmarks and three real-robot task families establishes a strong baseline.

This division of labor does not imply independent benefits. Across both ablation subsets, persistent memory and steering reinforce each other, while historical visual evidence complements semantic records. Richer language alone does not recover spatial guidance's benefit under the evaluated training setup.

Three limitations remain. First, to maintain the expert-data setting of benchmark baselines, we train on the same demonstrations rather than histories containing the Agent's own memory errors. Missing keyframes or incorrect facts can affect later decisions; their accumulation and recovery have not been systematically quantified. Second, language and point targets have limited expressiveness; traces, precise timing, or force constraints could support richer guidance. Third, supervision relies on fixed JSON schemas, and evaluation covers multi-task learning and within-task generalization; transfer to unseen tasks remains unverified.

Looking ahead, we envision flexible memory--action interfaces that adapt with experience, allowing robots to revise what they retain and how they convey it as tasks evolve.

\section*{LLM Usage Statement}
OpenAI ChatGPT and Codex assisted with research ideation, literature discovery, manuscript drafting and revision, pre-submission feedback, and code and LaTeX preparation. OpenAI image-generation tools assisted with illustrative figure assets, which the authors subsequently edited and assembled; they were not used to generate experimental observations or real-robot photographs. OpenAI Codex assisted with subtask segmentation and target-specific language relabeling for the expanded-language ablation. The authors take responsibility for the scientific claims, experimental results, references, and final presentation.

{\small
\bibliographystyle{ieeenat_fullname}
\bibliography{main}
}

\clearpage
\onecolumn
\raggedbottom
\setlength{\textfloatsep}{10pt plus 2pt minus 1pt}
\setlength{\intextsep}{8pt plus 2pt minus 1pt}
\setlength{\floatsep}{8pt plus 1pt minus 1pt}
\setlength{\dbltextfloatsep}{10pt plus 2pt minus 1pt}
\setlength{\dblfloatsep}{8pt plus 1pt minus 1pt}
\setcounter{topnumber}{4}
\setcounter{bottomnumber}{3}
\setcounter{totalnumber}{6}
\renewcommand{\topfraction}{0.9}
\renewcommand{\bottomfraction}{0.8}
\renewcommand{\textfraction}{0.08}
\renewcommand{\floatpagefraction}{0.65}
\makeatletter
\setlength{\@fptop}{0pt}
\setlength{\@fpsep}{10pt}
\setlength{\@fpbot}{0pt plus 1fil}
\renewcommand{\paragraph}{\@startsection{paragraph}{4}{\z@}{6pt plus 1pt minus 1pt}{-1em}{\normalfont\normalsize\bfseries}}
\makeatother
\appendix
\newcounter{suppfigure}
\renewcommand{\thesuppfigure}{S\arabic{suppfigure}}
\pdfbookmark[0]{Appendix}{appendix-title}
\begin{center}
\Large\bfseries Appendix
\end{center}
\section{Data Construction and Supervision}
\label{app:data}

This appendix summarizes data annotation, training, runtime and evaluation.

\subsection{Demonstrations and data splits}
\label{app:data-splits}
For each simulation benchmark, training uses all official expert demonstrations without a held-out validation split. Models are trained separately for each benchmark and evaluated through closed-loop simulator rollouts. Real-robot demonstrations are collected separately (Table~\ref{tab:app-real-data}).

\subsection{Subtask and keyframe annotation}
\label{app:annotation}
Subtask annotation is benchmark-specific (Table~\ref{tab:app-annotation}). For RoboDojo, metadata, state/action traces, gripper events and task events define a shared timeline for Agent and Action Model labels. Codex supplies subtask names, including return-to-position and button-confirmation stages. On LIBERO-Mem, the final holding stage is labeled ``Keep the arm still.''

Keyframe candidates include the initial frame, subtask transitions and grasp/release events. RMBench Press Button additionally uses local minima of the executing arm's end-effector height: the left arm for ordinary buttons and the right arm for confirmation. Each minimum is mapped to the nearest sampled window frame to refine keyframe labels while preserving subtask boundaries.
\begin{table*}[htbp]
\centering\small
\caption{Subtask annotation sources and dataset-specific refinements.}
\label{tab:app-annotation}
\begin{tabularx}{\textwidth}{@{}lXX@{}}
\toprule
Benchmark & Starting annotations & Refinements\\
\midrule
RoboMME & Official \texttt{simple\_subgoal\_online}, \texttt{grounded\_subgoal\_online} and \texttt{is\_demo}. & Scripted corrections to reference-phase causality, StopCube counts, and selected memory/keyframe labels.\\
RMBench & Official \texttt{language\_annotation} and operation segments. & Concise subgoal text, remaining-count descriptions and grounding; arm-specific height minima for Press Button keyframes.\\
RoboDojo & Metadata, state/action traces, gripper and task events; no official subgoals. & A shared timeline generates Agent/Action labels; Codex names subtasks. \\
LIBERO-Mem & Original subtask/goal and object annotations, together with grasp/place events. & A separate final holding stage, labeled ``Keep the arm still.''\\
\bottomrule
\end{tabularx}
\end{table*}

\subsection{Spatial steering labels}
\label{app:spatial-labels}
Steering labels use official grounded annotations, task-specific anchors and image annotations (Table~\ref{tab:app-steering}). LIBERO-Mem uses object bounding-box centers, with weighted mask centers as fallback. Swap T placement refers to the other T's initial position and orientation. These resources provide offline supervision; standard end-to-end evaluation uses observations and Agent-generated commands.

\begin{table*}[!htbp]
\centering\small
\caption{Sources of spatial steering supervision.}
\label{tab:app-steering}
\begin{tabularx}{\textwidth}{@{}lXX@{}}
\toprule
Benchmark & Main source & Target convention\\
\midrule
RoboMME & Points in official grounded subgoals. & Convert to the shared coordinate and role format.\\
RMBench & Official segments and task-specific initial visual/scene anchors. & Preserve target identity, original placement slots and T orientation where required.\\
RoboDojo & SAM segmentation and fixed slot points checked across episodes. & Use task-specific slot/target references.\\
LIBERO-Mem & Original object bounding boxes; original segmentation when boxes are unavailable. & Bounding-box center; weighted mask center as fallback. Empty masks produce missing labels.\\
\bottomrule
\end{tabularx}
\end{table*}

\subsection{Facts schemas and examples}
\label{app:facts}
Facts are lists of strings. The Agent input contains prior subgoal segments in \texttt{history}, a current segment in \texttt{active}, and a list in \texttt{facts}. When enabled, the output field \texttt{memory} replaces this list. Table~\ref{tab:app-facts-enabled} lists the relevant tasks, and Table~\ref{tab:app-facts-examples} gives examples.

Offline facts are generated by scripts from annotations, task events and initially visible layouts, with code-agent spot checks. Online, the VLM predicts the replacement list. Annotation preserves identity during movement and updates locations after an object settles. RMBench Press Button and RoboDojo Press by Number express remaining counts primarily in subgoal text, supported by history and event frames.

\begin{table*}[!htbp]
\centering\small
\caption{Tasks using the facts channel.}
\label{tab:app-facts-enabled}
\begin{tabularx}{\textwidth}{@{}lX@{}}
\toprule
Benchmark & Tasks\\
\midrule
RoboMME & ButtonUnmaskSwap, PatternLock, RouteStick, VideoPlaceButton, VideoPlaceOrder, VideoRepick, VideoUnmaskSwap, VideoUnmask.\\
RMBench & Cover Blocks, Swap Blocks, Swap T, Place Block Mat.\\
RoboDojo & Cover Blocks, Swap T, Imitate Sorting Sequence, Match and Pick from Conveyor.\\
LIBERO-Mem & No additional standalone facts channel.\\
\bottomrule
\end{tabularx}
\end{table*}

\begin{table*}[!htbp]
\centering\small
\caption{Examples of facts records from different task instances.}
\label{tab:app-facts-examples}
\begin{tabularx}{\textwidth}{@{}lX@{}}
\toprule
Task & Record text\\
\midrule
RMBench Cover Blocks & \texttt{Initial block locations: red right; green middle; blue left.}\\
RMBench Swap Blocks & \texttt{Initially occupied trays: left and right; middle empty.}\\
RoboDojo Swap T & \texttt{Red T initial anchor: XY [480, 575]; stem direction [-95, 31].}\\
RoboDojo sorting & \texttt{Demonstration order: doll -> action camera -> watch -> toy car -> mobile phone}\newline \texttt{Completed placements: doll -> action camera -> watch -> toy car -> mobile phone}\\
RoboMME PatternLock & \texttt{reference: path right -> right}\newline \texttt{execution: 1 direction segments completed}\\
RoboMME ButtonUnmaskSwap & \texttt{observed: container hiding blue cube now stationary at [642, 365]}\newline \texttt{execution: confirmed pickups 1; selected object released}\\
\bottomrule
\end{tabularx}
\end{table*}

\subsection{Causal training-sample construction}
\label{app:causal-sampling}
Current windows sample eight frames from the preceding execution, with strides determined by the action-chunk length. RMBench and RoboDojo use 25-step decision intervals. Training uses multiple offset lanes to cover different endpoints. Each lane assembles preceding memory, generates the current target, and then commits the update. Windows may cross subtask boundaries; the target follows the endpoint's subtask. Initialization uses the available episode prefix.

Inputs contain only preceding memory; current keyframe and facts updates become available at the next decision. Histories are teacher-forced during training and model-generated during deployment. Figure~\ref{fig:training} illustrates this causal ordering; Appendix~\ref{app:visual-runtime} describes online retention.

\section{Models and Supervised Training}
\label{app:implementation}
\label{app:training}

\subsection{Models and observation interfaces}
\label{app:models}
The Agent uses Qwen3-VL-4B-Instruct~\citep{bai2025qwen3vl}, with LoRA applied to linear layers in the language model, not the vision encoder. The Action Model uses a 3B PaliGemma backbone and a 300M action expert. Its image encoder is frozen and all remaining parameters are trained, without LoRA.

The Agent receives head/front-view images at $224\times224$ resolution. Table~\ref{tab:app-observations} summarizes Action Model observations. Each benchmark uses its own normalization statistics; LIBERO-Mem additionally converts end-effector coordinates and rotation representations.

\begin{table*}[!htbp]
\centering\small
\caption{Action Model observation interfaces. Simulation state vectors are padded to 32 dimensions.}
\label{tab:app-observations}
\begin{tabularx}{\textwidth}{@{}lXX@{}}
\toprule
Setting & Current camera views & State representation\\
\midrule
RoboMME~\citep{robomme2026} & Front and wrist views, following the official interface. & Seven joint positions and one gripper coordinate (8 dimensions).\\
RMBench / RoboDojo & Head and left/right wrists; $224\times224$ images. & Fourteen joint/gripper coordinates for two arms.\\
LIBERO-Mem & External and wrist views. & End-effector position, 6D rotation representation and gripper state.\\
Real robot & Head and left/right wrists; data recorded at 30 FPS. & Fourteen joint/gripper coordinates; absolute actions.\\
\bottomrule
\end{tabularx}
\end{table*}

\subsection{Agent fine-tuning}
\label{app:agent-training}
The Agent is trained for three epochs on each simulation benchmark by next-token prediction on serialized JSON targets, as in Eq.~\ref{eq:agent-sft}.
Table~\ref{tab:app-finetuning} compares the Agent and Action Model fine-tuning settings. The Agent uses eight GPUs, four examples per GPU and two gradient-accumulation steps, giving an effective batch of 64.

\begin{table}[htbp]
\centering\small
\caption{Fine-tuning settings for the Agent and Action Model.}
\label{tab:app-finetuning}
\label{tab:app-agent-config}
\label{tab:app-action-config}
\begin{tabularx}{\linewidth}{@{}lXX@{}}
\toprule
Parameter & Agent & Action Model\\
\midrule
Optimizer & AdamW & AdamW\\
Learning rate & $10^{-4}$ & $2.5\times10^{-5}$\\
Adam $\beta_1,\beta_2$ & $(0.9,0.95)$ & $(0.9,0.95)$\\
Adam $\epsilon$ & $10^{-8}$ & $10^{-8}$\\
Weight decay & $0.1$ & $10^{-10}$\\
Schedule & Cosine & Cosine decay to $10^{-6}$\\
Warm-up & $5\%$ of steps & Linear, first $10\%$ of steps\\
Gradient clipping & Maximum norm $1.0$ & Maximum norm $1.0$\\
Precision & BF16 & BF16\\
TF32 & Disabled & Enabled\\
Trainable parameters & Language-model LoRA adapters & All except the image encoder\\
LoRA rank; alpha; dropout & $16$; $32$; $0.05$ & Not used\\
Epochs & $3$ & $10$\\
\bottomrule
\end{tabularx}
\end{table}

\subsection{Action Model fine-tuning}
\label{app:action-training}
The Action Model is trained for ten epochs on each simulation benchmark using conditional flow matching~\citep{lipman2023flow}, with DeepSpeed ZeRO-1 and training seed 42. Let $V_t=\mathbf I_t$ denote current multiview images and $L_t$ the serialized command. With $b\sim\operatorname{Beta}(1.5,1)$, the implementation samples $\tau=0.999b+0.001$. An expert chunk $A_t^*$ and noise $\epsilon\sim\mathcal N(0,I)$ define $Z_\tau=\tau\epsilon+(1-\tau)A_t^*$. The target velocity is $\epsilon-A_t^*$, yielding
\begin{equation}
\mathcal L_{\mathrm{Action}}=\mathbb E_{t,\epsilon,\tau}\bigl[\|v_\phi(Z_\tau,\tau\mid V_t,q_t,L_t)-(\epsilon-A_t^*)\|_2^2\bigr].
\label{eq:action-fm}
\end{equation}
Inference starts from Gaussian noise and uses ten Euler steps toward the action endpoint (decreasing $\tau$). Tables~\ref{tab:app-finetuning} and~\ref{tab:app-action-chunks} list optimization settings, action spaces and chunk lengths.

\begin{table}[!htbp]
\centering\small
\caption{Action configurations. $H/h$ denotes predicted/executed steps.}
\label{tab:app-action-chunks}
\begin{tabularx}{\linewidth}{@{}lXr@{}}
\toprule
Benchmark & Action space & $H/h$\\
\midrule
RoboMME & Joint-position delta & $20/16$\\
RMBench & Absolute joints & $50/25$\\
RoboDojo & Absolute joints & $50/25$\\
LIBERO-Mem & EEF delta conversion & $16/16$\\
\bottomrule
\end{tabularx}
\end{table}

\section{Runtime Memory and Closed-Loop Execution}
\label{app:runtime}

\subsection{Prompt, output schema and command serialization}
\label{app:interface}
The Agent output maps $g_t,u_t,P_t,\widehat F_t$ to \texttt{task}, \texttt{hints}, \texttt{keyframe\_positions} and \texttt{memory}. Frame positions are one-based indices within the current window, not absolute episode indices. Point targets are integer image coordinates $[x,y]\in[0,1000]^2$. An optional \texttt{direction\_2d} is a signed relative vector of length approximately 100, not a second absolute point. A Swap T output is shown below.

\begingroup\footnotesize
\begin{verbatim}
{
 "task":
   "Use the left arm to place the red T-block.",
 "hints": [{
   "point_2d": [560, 590],
   "direction_2d": [-40, -91],
   "label": "place_target"
 }],
 "memory": [
   "Blue T initial anchor: XY [560, 590]; stem direction [-40, -91]."
 ],
 "keyframe_positions": [8]
}
\end{verbatim}
\endgroup
The Action Model receives the overall task and current subgoal in the following template:
\begingroup\footnotesize
\begin{verbatim}
Task: <global goal>;
Current Subgoal: {"task": "...", "hints": [...]};
Action:
\end{verbatim}
\endgroup
Neither \texttt{memory} nor \texttt{keyframe\_positions} enters this command. The original language-only interface retains only \texttt{task} in the subgoal object. Tasks without facts omit that input and its corresponding \texttt{memory} output.

\subsection{Visual memory maintenance}
\label{app:visual-runtime}
At each decision, eight frames are evenly sampled from the preceding execution, with strides adapted to the action-chunk length. Initialization uses the available prefix. Selected historical keyframes are combined with the current window in temporal order, removing duplicate indices. The full-memory condition also retains the initial frame; ablation inputs are specified in Appendix~\ref{app:ablations}.

\subsection{Subgoal history and facts maintenance}
\label{app:semantic-runtime}
The runtime initializes \texttt{history=[]}, \texttt{active=None} and \texttt{facts=[]}. To compare task descriptions, it strips surrounding whitespace, lowercases text and collapses repeated spaces. Consecutive equal tasks remain in one active segment; changing hint coordinates alone does not add a history entry. When the task or phase changes, the previous active segment is moved into history. These records describe issued decisions rather than verified successes.

For enabled tasks, \texttt{memory} replaces the complete facts list. An empty list clears facts; a missing required field or incorrect type causes parsing to fail.

\subsection{Parsing, timing and termination}
\label{app:execution}
Execution is synchronous, with Agent updates after each executed chunk (Table~\ref{tab:app-action-chunks}). Parsing failures trigger regeneration; only valid responses update memory or issue commands. The online feedback path does not expose simulator task state to the Agent.

\subsection{Rollout visualization}
\label{app:rollout-visualization}
\begin{figure}[!htbp]
\centering
\includegraphics[width=\linewidth]{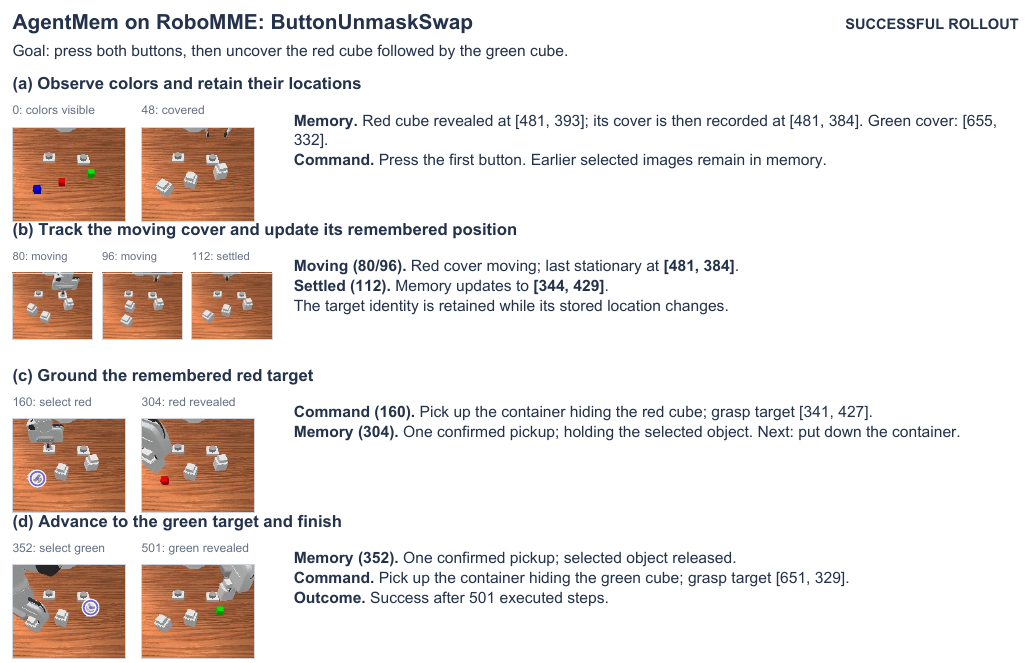}
\refstepcounter{suppfigure}\label{fig:app-mme-rollout}
\caption*{\textbf{Figure~\thesuppfigure. Successful RoboMME rollout.} Selected front-view crops and summaries of logged Agent outputs for ButtonUnmaskSwap. Indices denote executed control steps; circles mark predicted grasp targets in coordinates normalized to $[0,1000]^2$. The stored red-target position is updated after the cover moves. The complete rollout is provided in the supplementary video.}
\end{figure}

\begin{figure}[!htbp]
\centering
\includegraphics[width=\linewidth]{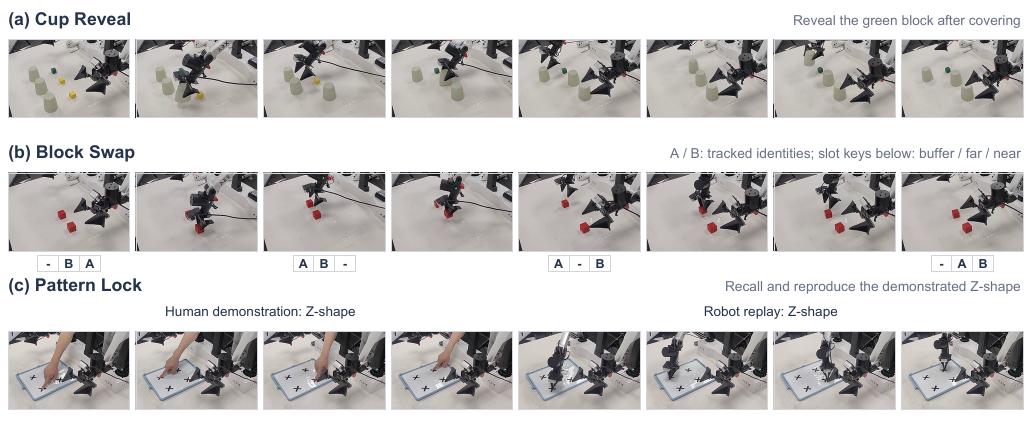}
\refstepcounter{suppfigure}\label{fig:app-real-rollouts}
\caption*{\textbf{Figure~\thesuppfigure. Third-person real-robot rollouts.} Each row shows one trial. Cup Reveal uncovers the green block. A/B track block identities; the external slot keys list buffer/far/near positions, with ``--'' denoting an empty slot. Pattern Lock pairs a human Z-shape demonstration with robot replay. Complete episodes are provided in the supplementary videos.}
\end{figure}

Figure~\ref{fig:app-mme-rollout} shows a successful RoboMME rollout with logged memory updates and grounded commands. Figure~\ref{fig:app-real-rollouts} shows one third-person rollout for each real-robot task. The supplementary videos provide the complete episodes and additional results, including Agent outputs and action execution. These videos will also be available on the project website upon public release.

\section{Simulation Evaluation and Ablation Protocols}
\label{app:evaluation}

\subsection{Task scope}
\label{app:sim-protocol}
The full-suite evaluation covers 16 RoboMME~\citep{robomme2026} tasks, ten RMBench~\citep{chen2026rmbench} tasks, ten LIBERO-Mem~\citep{liberomem2025} tasks and six RoboDojo Memory~\citep{robodojo2026} tasks. RMBench comparisons use the common nine tasks excluding Place-Mat, with ten-task means additionally reported where available.

\subsection{Metrics, grouping and aggregation}
\label{app:metrics}
RoboMME and RMBench use their official success standards. LIBERO-Mem strict SR requires the final official goal predicate, no overshoot and no runtime error. Completion is the fraction of subgoals confirmed in the official order. Its sequence evaluator includes sustained-satisfaction and leave-state counting conditions; breaking the final goal after completion can permanently set \texttt{\_overshot=True}. Relaxed SR permits overshoot. The original LIBERO-Mem baselines report completion rather than SR.

RoboDojo process scores are the task-specific per-episode \texttt{score} values in official task results. Its headline SR uses an equal-task macro average. The RMBench common-nine comparison excludes Place-Mat from the task set.
Across benchmarks, runs that terminate with runtime errors are counted as failures and remain in the trial denominator.

\subsection{Comparison scope}
\label{app:evidence-scope}

\paragraph{Reference selection and evaluation protocols.}
Other methods use the best reported configurations from original papers or benchmark reproductions under their source protocols, rather than matched retrainings. Appendix~\ref{app:comparison-scope} records relevant differences.

\paragraph{Benchmark coverage and training repeats.}
The four suites use benchmark-specific models of the same Agent--Action design. Full-suite results are fixed-run estimates; the six-task ablations report variation across three evaluation seeds (Appendix~\ref{app:ablations}).

\subsection{Trial accounting}
\label{app:exclusions}
RoboMME contributes 800 trials, RMBench 1,000, and LIBERO-Mem 1,000. RoboDojo contributes 300 trials, with 50 per task. Imitate sorting sequence has 0/50 successes. The total is 91/300 successes and task-macro SR remains 30.33\%.

The Imitate mean process score is 3.90 over 50 trials, and the six-task macro average is 31.73.

\subsection{Ablation configurations and controlled comparisons}
\label{app:ablations}
Table~\ref{tab:app-ablation-sharing} summarizes model sharing and retraining. Agents not shared with the original model are fine-tuned on three selected tasks per benchmark. Full-memory steering and original-language conditions share their Agent; the latter supplies only subgoal language to the Action Model. Semantic-only and current-window-only Agents share controller $S$. The current-window steering/language pair shares an Agent, and both original-language conditions share the three-task language-only controller $L$. Expanded language requires retraining both models on relabeled subgoals. This allocation ensures that each model is trained with its inference-time information types, with matched training budgets for each model type.

The current-window-only input removes historical images, subgoal history, the active decision and facts, including separate initial-frame retention. Training targets retain auxiliary memory fields, but their predicted outputs are not fed back at runtime.

\begin{table*}[!htbp]
\centering\small
\caption{Ablation training configurations within each benchmark. $S$ is the fixed steerable controller and $L$ the newly trained three-task language controller.}
\label{tab:app-ablation-sharing}
\begin{tabularx}{\textwidth}{@{}lXX@{}}
\toprule
Condition & Agent & Action Model\\
\midrule
Full memory + S & Original Agent. & Fixed $S$.\\
Semantic-only + S & Three-task SFT without historical images. & Same $S$.\\
Current window + S & Three-task SFT without persistent inputs. & Same $S$.\\
Current window + L & Same Agent as current window + S. & Newly trained three-task $L$.\\
Full memory + L & Same Agent as full memory + S. & Same $L$.\\
Full memory + L+ & Retrained on expanded subgoals. & Retrained on expanded subgoals.\\
Historical images + S & Three-task SFT to a MemER-like model. & Same $S$.\\
\bottomrule
\end{tabularx}
\end{table*}

\paragraph{Task selection.}
The subset covers complementary memory demands: trial-and-error ordering in Blocks Ranking, identity/location tracking in ButtonUnmaskSwap and Swap T, extended progress tracking in Press Button, and sequence-dependent decisions in PatternLock and VideoPlaceOrder. On Blocks Ranking, memory improves language-only SR from 72.7\% to 90.0\%.

\paragraph{Expanded-language control.}
Steering supplies target information as well as a spatial interface. To test whether more explicit language recovers its benefit, OpenAI Codex relabels subgoals using ground-truth targets and current images, requesting a unique reference within 128 tokens (Table~\ref{tab:app-expanded-examples}). Both models are retrained: the full-memory Agent generates expanded subgoals, and the language-only Action Model executes them. Across the six tasks in Table~\ref{tab:ablation}, expanded language improves mean SR by 1.67 points over original language but remains 42.44 points below steering.

\begin{table}[htbp]
\centering\small
\caption{Original and expanded subgoal templates. Placeholders specify the target identity, left-to-right object index or grid location.}
\label{tab:app-expanded-examples}
\begin{tabularx}{\linewidth}{@{}lXX@{}}
\toprule
Reference & Original subgoal & Expanded subgoal\\
\midrule
Object & Pick up the container that hides the \{color\} cube. & Lift container \{k\} of \{n\}, counting from left to right.\\
Motion & Move backward / left / forward-right. & Move the stick tip to the dot at row \{r\}, column \{c\}.\\
\bottomrule
\end{tabularx}
\end{table}

\paragraph{Grounding oracle descriptions.}
\label{app:language-grounding}
End-to-end success does not distinguish incorrect Agent descriptions from failures to execute a correct description. We therefore use RoboMME's online oracle to compare original subgoal language, expanded subgoal language, and subgoal language with 2D targets. For this diagnostic, expanded descriptions are generated by mapping oracle targets to language while simulator stepping is paused.

\begin{table}[htbp]
\centering\small
\caption{Oracle-command diagnostics on three RoboMME tasks. SR (\%); the mean weights tasks equally. The grounded condition records 43/50, 49/50 and 45/50 successes, respectively. These results are separate from the three-seed end-to-end ablations in Table~\ref{tab:ablation}.}
\label{tab:app-oracle-commands}
\begin{tabularx}{\linewidth}{@{}Xrrr@{}}
\toprule
Task & \shortstack{Oracle\\subgoal} & \shortstack{Oracle expanded\\subgoal} & \shortstack{Oracle subgoal\\+ 2D targets}\\
\midrule
ButtonUnmaskSwap & 16.00 & 44.00 & \textbf{86.00}\\
VideoPlaceOrder & 26.00 & 86.00 & \textbf{98.00}\\
PatternLock & \textbf{96.00} & 16.00 & 90.00\\
\midrule
Three-task mean & 46.00 & 48.67 & \textbf{91.33}\\
\bottomrule
\end{tabularx}
\end{table}

Oracle spatial guidance yields the highest three-task mean (91.33\%), compared with 46.00\% for original language and 48.67\% for expanded language. Expanded descriptions improve ButtonUnmaskSwap and VideoPlaceOrder, whereas concise directional language performs best on PatternLock. These oracle results show interface-dependent execution in the evaluated controllers, consistent with Steerable Policies~\citep{chen2026steerable}, which finds that geometric hints improve fine-grained control even through a textual interface.

The end-to-end ablations in Table~\ref{tab:ablation} use three evaluation seeds with 50 trials per task per seed on ButtonUnmaskSwap, VideoPlaceOrder, PatternLock, Blocks Ranking, Swap T and Press Button. Each configuration has 150 trials per task and 900 across the six tasks. The table reports equal-task subset means, not full-suite performance.

\section{Real-Robot Evaluation Details}
\label{app:real-robot}

\subsection{Hardware, observations and baselines}
\label{app:real-setup}
Real-robot deployment and evaluation use EVA-Client~\citep{yang2026evaclient}. The platform uses ARX X5 hardware with a head camera and left/right wrist cameras. Demonstrations are recorded at 30 FPS. The state contains 14 joint/gripper coordinates for the two arms, and actions are absolute rather than deltas. 
The study compares \method with single-frame and eight-frame $\pi_{0.5}$~\citep{black2025pi05}. All methods use the same number of expert demonstrations per task and the same Action Model training budget. The eight-frame baseline is trained and evaluated with visual history. Table~\ref{tab:app-real-data} lists the demonstration counts.

\begin{table}[!htbp]
\centering\small
\begin{minipage}[t]{0.33\linewidth}
\vspace{0pt}
\centering
\caption{Expert demonstrations per method.}
\label{tab:app-real-data}
\begin{tabular}{@{}lr@{}}
\toprule
Task & Demonstrations\\
\midrule
Cup Reveal & 15\\
Block Swap & 15\\
Pattern Lock & 80\\
\midrule
Total & 110\\
\bottomrule
\end{tabular}
\end{minipage}\hfill
\begin{minipage}[t]{0.64\linewidth}
\vspace{0pt}
\centering
\caption{Initial layouts. Cup targets are green; A/B denote tracked block identities.}
\label{tab:app-real-layouts}
\begin{tabular}{@{}lllll@{}}
\toprule
Case & Left & Middle & Right & Target\\
\midrule
Cup C1 & Green & Yellow & Yellow & Left\\
Cup C2 & Yellow & Green & Yellow & Middle\\
Cup C3 & Yellow & Yellow & Green & Right\\
\midrule
Swap S1 & Empty & A & B & Exchange A/B\\
Swap S2 & A & Empty & B & Exchange A/B\\
Swap S3 & A & B & Empty & Exchange A/B\\
\bottomrule
\end{tabular}
\end{minipage}
\end{table}

\subsection{Task configurations}
\label{app:real-tasks}

\paragraph{Cup Reveal.} The robot recalls the differently colored block before occlusion and uncovers it. Three layouts place the green block on the left, in the middle or on the right, with yellow blocks elsewhere (Table~\ref{tab:app-real-layouts}). Each layout has five expert demonstrations and five evaluation trials per method. Demonstrations allow natural placement variation within each nominal layout.
\paragraph{Block Swap.} The robot exchanges objects A and B through the empty buffer slot. The three layouts vary the empty slot (Table~\ref{tab:app-real-layouts}); each has five expert demonstrations and five evaluation trials per method.

\paragraph{Pattern Lock.} The robot reproduces a demonstrated sequence through four marked locations. Each of 16 non-crossing patterns is evaluated once per method. All test sequences appeared during training; a human demonstrates the target sequence anew for every evaluation episode. The Z and C examples in Figure~\ref{fig:real-robot-tasks} share locations but differ in order. Each method has 46 episodes across the three tasks.

\begin{figure}[!htbp]
\centering
\includegraphics[width=\linewidth]{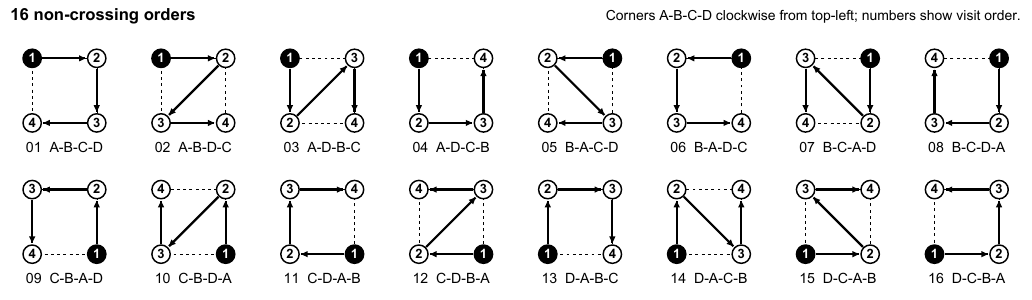}
\refstepcounter{suppfigure}\label{fig:app-four-corner-paths}
\caption*{\textbf{Figure~\thesuppfigure. Non-crossing four-corner paths.} Each open path visits all four corners once. Corners A--D run clockwise from the top-left; arrows and node numbers indicate visitation order, with a filled starting node. Reversed orders are distinct, giving 16 paths.}
\end{figure}

Figure~\ref{fig:app-four-corner-paths} illustrates the 16 directed, non-crossing open paths that visit each corner of a square once.

\subsection{Success and subgoal-completion scoring}
\label{app:real-scoring}
Completion credits each correctly placed cup and lifting the target cup in Cup Reveal; each completed pickup-and-placement transfer in Block Swap; and each point touched in the demonstrated order in Pattern Lock. Task SR is the fraction of successful episodes; completion is the fraction of prescribed subgoals achieved across episodes. Each stage contributes $25\%$ for Cup Reveal or Pattern Lock and one third for Block Swap (Table~\ref{tab:app-real-stages}). Pattern Lock achieves 60/64 subgoals (93.75\%), reported as 93.8\% in Table~\ref{tab:real}.

\begin{table*}[!htbp]
\centering\small
\caption{Stages credited for real-robot completion. A Block Swap stage requires both pickup and correct placement.}
\label{tab:app-real-stages}
\begin{tabularx}{\textwidth}{@{}lXr@{}}
\toprule
Task & Ordered stages & Credit per stage\\
\midrule
Cup Reveal & Place a cup over the left block; then the middle block; then the right block; lift the cup covering the differently colored target. & $25\%$\\
Block Swap & Move A to the empty slot; move B to A's original slot; move A to B's original slot. & $33\%$\\
Pattern Lock & Touch each of the four target points in the demonstrated order. & $25\%$\\
\bottomrule
\end{tabularx}
\end{table*}

\section{Additional Analyses and Reproducibility}
\label{app:analysis}

\subsection{Inference efficiency}
\label{app:efficiency}
\paragraph{Agent response latency on RoboMME.}
Timing spans request submission to the complete Agent response, excluding Action Model inference and robot execution. Under the same local RoboMME configuration on an A100 80GB, MemER takes 0.707\,s and adding semantic records increases this to 1.016\,s (43.7\%). The records add 128 input tokens on average, or 14.2\% of the visual-token count, approximately the token budget of two images. With caching and compiled inference, \method takes 0.812\,s on one A100 80GB and 0.663\,s on two.

\paragraph{Reported latency references.}
The original MemER report~\citep{sridhar2025memer}, Appendix D, Table 6, gives 0.787 $\pm$ 0.066\,s for Qwen2.5-VL-7B with eight recent frames and eight keyframes on an RTX 4090 (20 trials). Embodied-Astra~\citep{jedlicki2026embodiedastra} reports 18.3\,s per call on RoboMME, with Cartesian waypoints executed by a motion planner. Astranomy~\citep{yang2026astranomy} reports 18.56\,s per direct-policy call on RoboLab-120 over successful episodes, including network delay. These cross-system call times use different hardware and interfaces and are not end-to-end speed comparisons.

\subsection{Comparison with agent-side memory systems}
\label{app:provenance}
\paragraph{2AM.}
In its original zero-shot setting, 2AM~\citep{hu2026twoam} supplies history through an off-the-shelf Agent's context and emphasizes robust VLA training with condition dropout, spatial noise and temporal jitter. \method supervises recurrent Agent-side memory updates and grounded decisions, and evaluates memory--steering interactions across four benchmarks. On LIBERO-Mem, \method achieves 54.10\% strict SR, compared with 11.83\% reported for the original zero-shot 2AM.

\paragraph{MEM.}
MEM~\citep{torne2026mem} trains semantic-memory updates from external-LLM summaries and combines them with short-term video memory for control, evaluated on custom real-robot tasks. \method provides a strong baseline on four public memory benchmarks with retrained memory/interface controls. Full training, annotation and evaluation code and configurations will be released to support direct comparison.

\section{Fixed-Version Simulation Results}
\label{app:fixed-results}
\subsection{Per-task results}
\label{app:task-results}
These tables report full-suite per-task results. RoboMME uses one model jointly trained on all 16 tasks. Aggregate rates weight tasks equally.
\begin{table*}[!htbp]
\centering\small
\setlength{\tabcolsep}{5pt}
\renewcommand{\arraystretch}{1.06}
\caption{RoboMME per-task results (\%); 50 trials per task.}
\label{tab:tasks-robomme}
\begin{tabularx}{\textwidth}{@{}Xlrr@{}}
\toprule Task & Type & Successes / trials & SR \\ \midrule
BinFill & Counting & 40/50 & 80.00 \\
PickXtimes & Counting & 50/50 & 100.00 \\
SwingXtimes & Counting & 48/50 & 96.00 \\
StopCube & Counting & 22/50 & 44.00 \\
VideoUnmask & Permanence & 50/50 & 100.00 \\
ButtonUnmask & Permanence & 47/50 & 94.00 \\
VideoUnmaskSwap & Permanence & 32/50 & 64.00 \\
ButtonUnmaskSwap & Permanence & 32/50 & 64.00 \\
PickHighlight & Reference & 39/50 & 78.00 \\
VideoRepick & Reference & 22/50 & 44.00 \\
VideoPlaceButton & Reference & 41/50 & 82.00 \\
VideoPlaceOrder & Reference & 39/50 & 78.00 \\
MoveCube & Imitation & 39/50 & 78.00 \\
InsertPeg & Imitation & 6/50 & 12.00 \\
PatternLock & Imitation & 30/50 & 60.00 \\
RouteStick & Imitation & 24/50 & 48.00 \\
\midrule Macro average & & 561/800 & \textbf{70.13} \\
\bottomrule\end{tabularx}
\end{table*}

\begin{table*}[!htbp]
\centering\small
\setlength{\tabcolsep}{5pt}
\renewcommand{\arraystretch}{1.06}
\caption{RMBench per-task results (\%); 100 trials per task.}
\label{tab:tasks-rmbench}
\begin{tabularx}{\textwidth}{@{}Xlrr@{}}
\toprule Task & Type & Successes / trials & SR \\ \midrule
observe\_and\_pickup & M(1) & 47/100 & 47.00 \\
rearrange\_blocks & M(1) & 60/100 & 60.00 \\
put\_back\_block & M(1) & 98/100 & 98.00 \\
swap\_blocks & M(1) & 59/100 & 59.00 \\
swap\_T & M(1) & 76/100 & 76.00 \\
battery\_try & M(n) & 35/100 & 35.00 \\
blocks\_ranking\_try & M(n) & 91/100 & 91.00 \\
cover\_blocks & M(n) & 65/100 & 65.00 \\
press\_button & M(n) & 30/100 & 30.00 \\
place\_block\_mat & Additional task & 66/100 & 66.00 \\
\midrule Macro average & & 627/1000 & \textbf{62.70} \\
Common nine tasks (excluding Place-Mat) & & 561/900 & \textbf{62.33} \\
\bottomrule\end{tabularx}
\end{table*}

\begin{table*}[!htbp]
\centering\small
\setlength{\tabcolsep}{5pt}
\renewcommand{\arraystretch}{1.06}
\caption{LIBERO-Mem fixed-version task results (\%). Strict SR is separate from relaxed SR and completion.}
\label{tab:tasks-libero-mem}
\begin{tabularx}{\textwidth}{@{}lXlrrr@{}}
\toprule Task & Description & Type & Strict SR & Relaxed SR & Completion \\ \midrule
T1 & Lift bowl and return to plate (1x) & OM only & 75.00 & 75.00 & 75.00 \\
T2 & Lift bottle and return to plate (1x) & OM only & 81.00 & 86.00 & 86.00 \\
T3 & Lift bowl and return to plate (3x) & OM+OS & 19.00 & 23.00 & 60.33 \\
T4 & Lift bottle and return to plate (3x) & OM+OS & 49.00 & 53.00 & 81.67 \\
T5 & Lift bowl and return to plate (5x) & OM+OS & 24.00 & 26.00 & 68.00 \\
T6 & Lift bowl and return to plate (7x) & OM+OS & 5.00 & 5.00 & 56.43 \\
T7 & Swap two bowls using the empty plate & OM+OR & 62.00 & 63.00 & 82.00 \\
T8 & Rotate three bowls using the empty plate & OM+OR & 59.00 & 60.00 & 78.75 \\
T9 & Put cheese in nearest basket; move that basket & OM+OO & 69.00 & 69.00 & 83.00 \\
T10 & Put cheese in nearest basket; move empty basket & OM+OO & 98.00 & 99.00 & 99.50 \\
\midrule \multicolumn{3}{l}{Macro average (1,000 trials)} & \textbf{54.10} & \textbf{55.90} & \textbf{77.07} \\
\bottomrule\end{tabularx}
\end{table*}

\begin{table*}[!htbp]
\centering\small
\setlength{\tabcolsep}{5pt}
\renewcommand{\arraystretch}{1.06}
\caption{RoboDojo Memory fixed-version task results (\%). Each task uses 50 trials. Process scores are averaged over all trials.}
\label{tab:tasks-robodojo-mem}
\begin{tabularx}{\textwidth}{@{}Xrrr@{}}
\toprule Task & Successes / trials & SR & Process score \\ \midrule
cover\_blocks & 35/50 & 70.00 & 74.50 \\
swap\_blocks & 3/50 & 6.00 & 6.00 \\
swap\_T & 2/50 & 4.00 & 4.00 \\
press\_by\_number & 24/50 & 48.00 & 48.00 \\
imitate\_sorting\_sequence & 0/50 & 0.00 & 3.90 \\
match\_and\_pick\_from\_conveyor & 27/50 & 54.00 & 54.00 \\
\midrule Macro average & 300 trials & \textbf{30.33} & 31.73 \\
\bottomrule\end{tabularx}
\end{table*}

\subsection{Figure construction and reference provenance}
\label{app:figure-protocol}
\paragraph{Reference provenance and inclusion.}
Figure~\ref{fig:main-results} compares benchmark-native baselines and recent methods. Numerical sources and protocol notes are collected in Appendix~\ref{app:external-comparisons}. RoboMME~\citep{robomme2026} supplies its reference scores; RMBench~\citep{chen2026rmbench} uses its Table~I and EventVLA~\citep{yang2026eventvla} Table~9. LIBERO-Mem~\citep{liberomem2025} supplies completion scores, including SlotSSM with oracle subgoals. RoboDojo~\citep{robodojo2026} SRs come from task logs or the cited reports.
\paragraph{Plot construction.} A group's bar width is $(n_c/N)r_c$, where $r_c$ is its task-macro rate and $n_c/N$ its task weight; adjacent tables show the raw $r_c$. LIBERO-Mem pairs SR with completion. Gray bars indicate unavailable subgroup breakdowns, and N/R marks unreported metrics. Source-reported totals are retained; rounded task entries may yield slightly different means.

\begingroup
\makeatletter
\setlength{\@fptop}{0pt}
\setlength{\@fpsep}{12pt}
\setlength{\@fpbot}{0pt plus 1fil}
\makeatother
\section{Extended reported comparisons}
\label{app:external-comparisons}
\subsection{Comparison scope and protocol differences}
\label{app:comparison-scope}
Figure~\ref{fig:main-results} and the following tables compare benchmark-native baselines and recent methods. Sources were collected on 9 November 2026. Source codes identify provenance; ``--'' denotes unreported values. Human and oracle references are labeled separately.
\paragraph{Evaluation settings.} RoboDojo results use the official task-log snapshot of 3 November. GPT-6 Astra, GPT-5.5 and DeepSeek-Flash SRs~\citep{zhang2026unexpected} come from those logs; DeepSeek uses 10 episodes per task and the GPT policies use 50. Embodied-Astra~\citep{jedlicki2026embodiedastra} uses a waypoint planner. HexaShell v3 uses 30 trials per task on public layouts with calibrated geometry and motion tools. Agent as Policy reports a seed-0 evaluation; Astra-Guide uses cross-episode experience.
\paragraph{Metric conventions.} LIBERO-Mem SR and completion are reported separately. External SR is source-reported; parity with the strict evaluator used by \method is not assumed. SmoLSTM~\citep{habekost2026smolstm} uses a relaxed-SR reference and calls completion ``coverage.'' Its flow variant uses 20 trials per task; MemoryVAM uses the UNet variant. TRP~\citep{aktas2026trp} reports completion. RMBench columns distinguish the common nine-task and complete ten-task means. Where task-log means and website summaries differ, the listed task-log values are retained.
\subsection{Source inventory and per-task comparisons}
\label{app:comparison-sources}
\paragraph{Source key.}
\textbf{MM}: \href{https://robomme.github.io/leaderboard.html}{RoboMME official leaderboard}.
\textbf{DJ}: \href{https://robodojo-benchmark.com/leaderboard}{RoboDojo official task logs, 3 November snapshot}.
\textbf{EV}: \href{https://arxiv.org/html/2606.20092}{EventVLA, Appendix C, Table 9; RoboMME aggregate only}.
\textbf{SM}: \href{https://arxiv.org/html/2609.05533}{SimpleMemVLA, Tables 1 and 10}.
\textbf{LT}: \href{https://arxiv.org/html/2610.21362}{LTM, Tables 1 and 2}.
\textbf{EM}: \href{https://arxiv.org/html/2610.05418}{EvoMem-VLA, Tables 1 and 2}.
\textbf{MA}: \href{https://arxiv.org/html/2610.22323}{Mem4Act, Tables 1 and 2}.
\textbf{T2}: \href{https://arxiv.org/html/2609.36720}{T2Mem, per-task RoboMME table}.
\textbf{EC}: \href{https://arxiv.org/html/2610.00801}{ECoMEM, Table 1}.
\textbf{SE}: \href{https://arxiv.org/html/2610.29560}{SERA, Appendix A, Table 2}.
\textbf{HX}: \href{https://arxiv.org/html/2610.23479v3}{HexaShell v3, Tables 3 and 10; public layouts, 30 trials/task}.
\textbf{AP}: \href{https://huggingface.co/datasets/Agent-as-Policy/agent-as-policy-robodojo}{Agent as Policy, released seed-0 evaluation, 50 trials/task}.
\textbf{AG}: \href{https://legend717.github.io/astra-guide-vla-website/report/}{Astra-Guide technical report; cross-episode experience}.
\textbf{EA}: \href{https://arxiv.org/abs/2610.17350}{Embodied-Astra; waypoint policy, Table 1}.
\textbf{UP}: \href{https://arxiv.org/abs/2609.24170}{Unexpected Robot Policy; Memory SR in Table 5; per-task SR from DJ, not the process scores of Table 6}.
\textbf{PP}: \href{https://arxiv.org/abs/2608.24115}{PonderPounce}.
\textbf{LM}: \href{https://ojs.aaai.org/index.php/AAAI/article/view/37337}{LIBERO-Mem, Table 4: completion, not SR}.
\textbf{AgentMem}: Results reported in Appendix~\ref{app:fixed-results}.
\textbf{SL}: \href{https://arxiv.org/html/2609.22854}{SmoLSTM, Table I and Section IV-A}.
\textbf{TR}: \href{https://arxiv.org/html/2610.33288}{Tiny Recursive Policy, Section 4.3 and Table 2}.
\textbf{MV}: \href{https://arxiv.org/html/2606.20679}{MemoryVAM, original paper}.
\begin{table*}[!htbp]
\centering\scriptsize
\setlength{\tabcolsep}{3pt}
\caption{RoboMME success rates, tasks 1--8 (rows 1--24). BF: BinFill, PX: PickXtimes, SX: SwingXtimes, SC: StopCube, VU: VideoUnmask, BU: ButtonUnmask, VUS: VideoUnmaskSwap, BUS: ButtonUnmaskSwap.}
\begin{tabular}{@{}p{4.1cm}lrrrrrrrrr@{}}
\toprule
Method & Source & BF & PX & SX & SC & VU & BU & VUS & BUS & 16-task Avg. \\
\midrule
MEMI-Astra-Plan (oracle planner) & MM & 98 & 100 & 100 & 100 & 100 & 98 & 98 & 100 & 99.38 \\
Astra-Guide & MM & 96 & 100 & 96 & 96 & 100 & 100 & 98 & 92 & 97.13 \\
MEMI-Astra-Act & MM & 98 & 100 & 100 & 96 & 100 & 100 & 96 & 92 & 97 \\
Embodied-Astra & MM,EA & 98 & 100 & 98 & 54 & 94 & 100 & 96 & 98 & 92.13 \\
Human Performance & MM & 96 & 100 & 80 & 78 & 90 & 92 & 92 & 90 & 90.5 \\
SimpleMemVLA & MM,SM & 78 & 100 & 96 & 92 & 100 & 100 & 90 & 94 & 88.25 \\
GroundSG+Oracle & MM & 85.78 & 100 & 100 & 49.67 & 98.78 & 95 & 99.22 & 80.22 & 84.08 \\
ECoMEM & EC & 71.33 & 100 & 99.33 & 66 & 100 & 96.67 & 78 & 84 & 82.42 \\
EvoMem-VLA & EM & -- & -- & -- & -- & -- & -- & -- & -- & 82 \\
3-tier Astra & MM & 82 & 96 & 82 & 18 & 92 & 100 & 100 & 84 & 79.13 \\
\textbf{AgentMem (current)} & \textbf{AgentMem} & \textbf{80} & \textbf{100} & \textbf{96} & \textbf{44} & \textbf{100} & \textbf{94} & \textbf{64} & \textbf{64} & \textbf{70.13} \\
PonderPounce & MM,LT,PP & 50.67 & 93.33 & 72 & 82.67 & 94.67 & 97.33 & 25.33 & 34 & 60.83 \\
LTM-pi (DreamDojo) & LT & 66 & 99.33 & 96 & 100 & 72.67 & 86.67 & 34 & 36 & 58.54 \\
LTM-pi (CD-LAM) & LT & 64.67 & 100 & 93.33 & 87.33 & 68.67 & 98 & 30.67 & 31.33 & 57.67 \\
T2Mem & MM,T2 & 60.67 & 91.33 & 90.67 & 70 & 88 & 59.33 & 29.33 & 20 & 56.83 \\
SERA & SE & 48 & 70 & 72 & 6 & 96 & 40 & 92 & 0 & 51.38 \\
Mem4Act & MA & -- & -- & -- & -- & -- & -- & -- & -- & 49.88 \\
SimpleSG+Oracle & MM & 85.78 & 99.78 & 100 & 44.67 & 33.11 & 22 & 15.56 & 15.56 & 49.58 \\
MemER (SERA reproduction) & SE & 60 & 82 & 72 & 4 & 86 & 82 & 36 & 24 & 45.38 \\
FrameSamp+Modul & MM & 39.56 & 87.33 & 92 & 42 & 32.67 & 25.11 & 24.44 & 18.22 & 44.51 \\
MemER & MM & 56.67 & 79.33 & 59.33 & 0 & 81.33 & 72 & 38 & 21.33 & 42.38 \\
TokenDrop+Modul & MM & 34.44 & 83.56 & 86 & 5.33 & 28.22 & 29.33 & 28.44 & 21.33 & 38.04 \\
FrameSamp+Expert & MM & 57.33 & 86.22 & 94.67 & 28.89 & 31.78 & 25.78 & 22.89 & 20.22 & 36.25 \\
TokenDrop+Expert & MM & 54.22 & 87.56 & 91.78 & 4.22 & 26.67 & 30.44 & 18.44 & 18.89 & 34.86 \\
\bottomrule
\end{tabular}
\end{table*}
\begin{table*}[!htbp]
\centering\scriptsize
\setlength{\tabcolsep}{3pt}
\caption{RoboMME success rates, tasks 1--8 (rows 25--47). BF: BinFill, PX: PickXtimes, SX: SwingXtimes, SC: StopCube, VU: VideoUnmask, BU: ButtonUnmask, VUS: VideoUnmaskSwap, BUS: ButtonUnmaskSwap.}
\begin{tabular}{@{}p{4.1cm}lrrrrrrrrr@{}}
\toprule
Method & Source & BF & PX & SX & SC & VU & BU & VUS & BUS & 16-task Avg. \\
\midrule
TokenDrop+Context & MM & 48.67 & 85.11 & 94.67 & 3.11 & 33.78 & 31.56 & 26.22 & 16 & 34.5 \\
GroundSG+QwenVL & MM & 52 & 92.67 & 7.33 & 0 & 88.67 & 24 & 30.67 & 14 & 32.7 \\
EventVLA (VA + KEM) & EV & -- & -- & -- & -- & -- & -- & -- & -- & 32.1 \\
SimpleMemVLA (Retrieval) & SM & 38 & 34 & 76 & 38 & 76 & 58 & 20 & 14 & 31.5 \\
FrameSamp+Context & MM & 41.22 & 72 & 73.67 & 13.67 & 26.89 & 30.22 & 20.89 & 15.22 & 30.68 \\
SimpleSG+QwenVL & MM & 77.56 & 95.33 & 5.11 & 0.44 & 34.22 & 19.33 & 15.33 & 9.56 & 29 \\
SimpleSG+Gemini & MM & 46 & 63 & 45 & 2 & 29 & 9 & 14 & 2 & 23.25 \\
SimpleMemVLA (Token Compression) & SM & 26 & 50 & 92 & 16 & 18 & 12 & 14 & 10 & 22.63 \\
TTT+Expert & MM & 34.89 & 63.78 & 41.33 & 4 & 31.78 & 22.44 & 19.56 & 18 & 22.35 \\
TTT+Context & MM & 35.56 & 62.89 & 42.44 & 3.33 & 29.78 & 22.89 & 18.44 & 14.44 & 22.28 \\
TTT+Modul & MM & 34.22 & 65.11 & 36.67 & 2.11 & 27.22 & 22.11 & 25.22 & 14.11 & 21.96 \\
SAM2Act & MM & 40 & 76 & 25.33 & 0 & 27.33 & 32 & 18 & 26.67 & 21.37 \\
SimpleMemVLA (Recurrent State) & SM & 26 & 36 & 56 & 6 & 26 & 24 & 30 & 16 & 20.63 \\
RMT+Modul & MM & 33.33 & 60.78 & 37.78 & 4.67 & 31.11 & 11.78 & 17.78 & 2.44 & 20.17 \\
pi0.5 w/ action history & MM & 26.67 & 58.33 & 26.67 & 4.67 & 30.67 & 23.67 & 20.67 & 16 & 19.73 \\
RMT+Context & MM & 32.44 & 56.89 & 33.56 & 5.78 & 31.33 & 10.89 & 17.33 & 2 & 19.46 \\
RMT+Expert & MM & 35.78 & 60.22 & 36 & 5.56 & 28 & 17.11 & 15.78 & 2 & 18.15 \\
WeaveLA (oracle) & LT & 28 & 34 & 50 & 12 & 24 & 28 & 22 & 4 & 18.12 \\
pi0.5 & MM & 30 & 42.89 & 35.56 & 6.67 & 20.44 & 22.22 & 18.67 & 6.67 & 17.93 \\
WeaveLA & LT & 26 & 44 & 46 & 2 & 28 & 14 & 14 & 4 & 17.38 \\
GroundSG+Gemini & MM & 26 & 18 & 4 & 3 & 36 & 14 & 13 & 0 & 11.56 \\
MemoryVLA & LT & 10 & 17.33 & 1.33 & 0 & 16 & 8 & 4.67 & 4.67 & 7.12 \\
DP-Unet & MM & 32 & 2.67 & 6.67 & 0 & 16 & 4 & 4 & 0 & 7.11 \\
\bottomrule
\end{tabular}
\end{table*}
\begin{table*}[!htbp]
\centering\scriptsize
\setlength{\tabcolsep}{3pt}
\caption{RoboMME success rates, tasks 9--16 (rows 1--24). PH: PickHighlight, VR: VideoRepick, VPB: VideoPlaceButton, VPO: VideoPlaceOrder, MC: MoveCube, IP: InsertPeg, PL: PatternLock, RS: RouteStick.}
\begin{tabular}{@{}p{4.1cm}lrrrrrrrrr@{}}
\toprule
Method & Source & PH & VR & VPB & VPO & MC & IP & PL & RS & 16-task Avg. \\
\midrule
MEMI-Astra-Plan (oracle planner) & MM & 100 & 100 & 100 & 100 & 100 & 96 & 100 & 100 & 99.38 \\
Astra-Guide & MM & 100 & 100 & 98 & 90 & 100 & 90 & 100 & 98 & 97.13 \\
MEMI-Astra-Act & MM & 100 & 92 & 100 & 100 & 96 & 84 & 100 & 98 & 97 \\
Embodied-Astra & MM,EA & 98 & 100 & 100 & 98 & 54 & 92 & 100 & 94 & 92.13 \\
Human Performance & MM & 92 & 92 & 98 & 90 & 90 & 98 & 84 & 86 & 90.5 \\
SimpleMemVLA & MM,SM & 90 & 64 & 86 & 90 & 94 & 46 & 94 & 98 & 88.25 \\
GroundSG+Oracle & MM & 83.33 & 97.33 & 100 & 100 & 87.78 & 15.56 & 97 & 55.56 & 84.08 \\
ECoMEM & EC & 62.67 & 77.33 & 96 & 86 & 96 & 40.67 & 79.33 & 85.33 & 82.42 \\
EvoMem-VLA & EM & -- & -- & -- & -- & -- & -- & -- & -- & 82 \\
3-tier Astra & MM & 84 & 86 & 100 & 98 & 80 & 18 & 88 & 58 & 79.13 \\
\textbf{AgentMem (current)} & \textbf{AgentMem} & \textbf{78} & \textbf{44} & \textbf{82} & \textbf{78} & \textbf{78} & \textbf{12} & \textbf{60} & \textbf{48} & \textbf{70.13} \\
PonderPounce & MM,LT,PP & 85.33 & 40.67 & 83.33 & 79.33 & 81.33 & 6 & 12.67 & 34.67 & 60.83 \\
LTM-pi (DreamDojo) & LT & 38.67 & 42.67 & 38 & 30.67 & 77.33 & 20 & 60 & 38.67 & 58.54 \\
LTM-pi (CD-LAM) & LT & 22.67 & 30 & 38.67 & 22.67 & 80 & 20.67 & 64 & 70 & 57.67 \\
T2Mem & MM,T2 & 50.67 & 38.67 & 35.33 & 30 & 80.67 & 36 & 50 & 78.67 & 56.83 \\
SERA & SE & 54 & 34 & 98 & 92 & 60 & 0 & 60 & 0 & 51.38 \\
Mem4Act & MA & -- & -- & -- & -- & -- & -- & -- & -- & 49.88 \\
SimpleSG+Oracle & MM & 44 & 27.78 & 31.33 & 26 & 87.33 & 10 & 95.33 & 55.11 & 49.58 \\
MemER (SERA reproduction) & SE & 70 & 24 & 28 & 36 & 90 & 6 & 20 & 6 & 45.38 \\
FrameSamp+Modul & MM & 22.89 & 30.44 & 60 & 32 & 77.78 & 7.56 & 53.56 & 66.67 & 44.51 \\
MemER & MM & 70.67 & 25.33 & 30 & 26 & 82.67 & 6.67 & 16.67 & 12 & 42.38 \\
TokenDrop+Modul & MM & 21.33 & 22 & 59.56 & 36 & 62 & 7.11 & 32.44 & 51.56 & 38.04 \\
FrameSamp+Expert & MM & 19.11 & 23.11 & 30 & 24.22 & 83.11 & 2 & 13.56 & 17.11 & 36.25 \\
TokenDrop+Expert & MM & 19.33 & 20.89 & 36.44 & 24.67 & 87.56 & 2.22 & 16.22 & 18.22 & 34.86 \\
\bottomrule
\end{tabular}
\end{table*}
\begin{table*}[!htbp]
\centering\scriptsize
\setlength{\tabcolsep}{3pt}
\caption{RoboMME success rates, tasks 9--16 (rows 25--47). PH: PickHighlight, VR: VideoRepick, VPB: VideoPlaceButton, VPO: VideoPlaceOrder, MC: MoveCube, IP: InsertPeg, PL: PatternLock, RS: RouteStick.}
\begin{tabular}{@{}p{4.1cm}lrrrrrrrrr@{}}
\toprule
Method & Source & PH & VR & VPB & VPO & MC & IP & PL & RS & 16-task Avg. \\
\midrule
TokenDrop+Context & MM & 20.67 & 17.78 & 31.11 & 25.33 & 81.33 & 4 & 12.67 & 20 & 34.5 \\
GroundSG+QwenVL & MM & 15.11 & 25.33 & 54 & 31.78 & 71.56 & 3.33 & 6.67 & 6 & 32.7 \\
EventVLA (VA + KEM) & EV & -- & -- & -- & -- & -- & -- & -- & -- & 32.1 \\
SimpleMemVLA (Retrieval) & SM & 18 & 0 & 18 & 18 & 68 & 8 & 10 & 10 & 31.5 \\
FrameSamp+Context & MM & 17.67 & 15.22 & 30 & 20.89 & 77.22 & 1.22 & 15.22 & 19.67 & 30.68 \\
SimpleSG+QwenVL & MM & 17.11 & 25.33 & 33.33 & 25.11 & 82 & 3.78 & 12.67 & 7.78 & 29 \\
SimpleSG+Gemini & MM & 20 & 15 & 26 & 29 & 61 & 4 & 7 & 0 & 23.25 \\
SimpleMemVLA (Token Compression) & SM & 16 & 6 & 16 & 12 & 56 & 0 & 10 & 8 & 22.63 \\
TTT+Expert & MM & 12.22 & 9.56 & 34 & 22.89 & 33.56 & 0.89 & 3.11 & 5.56 & 22.35 \\
TTT+Context & MM & 20.44 & 13.11 & 34.22 & 20.22 & 32.44 & 1.11 & 1.56 & 3.56 & 22.28 \\
TTT+Modul & MM & 14.56 & 12.11 & 32.67 & 22.33 & 31.22 & 1.11 & 3.56 & 7 & 21.96 \\
SAM2Act & MM & 17.33 & 5.33 & 24.67 & 20 & 29.33 & 0 & 0 & 0 & 21.37 \\
SimpleMemVLA (Recurrent State) & SM & 12 & 4 & 34 & 20 & 16 & 0 & 10 & 14 & 20.63 \\
RMT+Modul & MM & 17.11 & 4.22 & 32 & 31.11 & 24.67 & 2.21 & 3.78 & 8 & 20.17 \\
pi0.5 w/ action history & MM & 12.33 & 8.67 & 24 & 18.67 & 34 & 1 & 4 & 5.67 & 19.73 \\
RMT+Context & MM & 14 & 3.78 & 32 & 29.11 & 25.78 & 2 & 5.56 & 8.89 & 19.46 \\
RMT+Expert & MM & 11.78 & 0.22 & 24.22 & 22.67 & 20 & 1.89 & 4.22 & 4.89 & 18.15 \\
WeaveLA (oracle) & LT & 12 & 2 & 22 & 14 & 16 & 2 & 8 & 12 & 18.12 \\
pi0.5 & MM & 11.33 & 0.44 & 31.11 & 25.78 & 26 & 1.56 & 2.89 & 4.67 & 17.93 \\
WeaveLA & LT & 10 & 4 & 26 & 18 & 22 & 4 & 8 & 8 & 17.38 \\
GroundSG+Gemini & MM & 9 & 17 & 12 & 7 & 17 & 0 & 7 & 2 & 11.56 \\
MemoryVLA & LT & 7.33 & 0 & 13.33 & 4 & 14 & 0 & 12 & 1.33 & 7.12 \\
DP-Unet & MM & 0 & 0 & 22 & 5.67 & 8.67 & 0 & 2 & 10 & 7.11 \\
\bottomrule
\end{tabular}
\end{table*}
\begin{table*}[!htbp]
\centering\scriptsize
\setlength{\tabcolsep}{3pt}
\caption{LIBERO-Mem reported success rates (rows 1--12).}
\begin{tabular}{@{}p{4.1cm}lrrrrrrrrrrr@{}}
\toprule
Method & Source & T1 & T2 & T3 & T4 & T5 & T6 & T7 & T8 & T9 & T10 & Avg. \\
\midrule
LTM-pi (DreamDojo) & LT & 95 & 75 & 100 & 90 & 90 & 65 & 65 & 25 & 100 & 100 & 80.5 \\
SmoLSTM (flow) & SL & -- & -- & -- & -- & -- & -- & -- & -- & -- & -- & 77.5 \\
LTM-pi (CD-LAM) & LT & 95 & 75 & 90 & 70 & 70 & 45 & 80 & 20 & 100 & 100 & 74.5 \\
\textbf{AgentMem (current)} & \textbf{AgentMem} & \textbf{75} & \textbf{81} & \textbf{19} & \textbf{49} & \textbf{24} & \textbf{5} & \textbf{62} & \textbf{59} & \textbf{69} & \textbf{98} & \textbf{54.1} \\
MemoryVAM (UNet) & LT,MV & 85 & 65 & 90 & 10 & 35 & 0 & 25 & 5 & 40 & 70 & 42.5 \\
GMP & LT & 85 & 45 & 20 & 5 & 40 & 0 & 0 & 0 & 0 & 0 & 19.5 \\
pi0 (2AM reimplementation) & LT & 30.83 & 13.33 & 14.17 & 21.67 & 11.67 & 10 & 6.67 & 8.33 & 0 & 5.83 & 12.25 \\
2AM (earlier workshop) & LT & 1.67 & 5 & 5 & 6.67 & 6.67 & 2.5 & 7.5 & 10 & 47.5 & 25.83 & 11.83 \\
MemoryVLA & LT & 55 & 0 & 35 & 0 & 5 & 0 & 0 & 0 & 0 & 0 & 9.5 \\
pi0.5 (original) & LT & 0 & 0 & 0 & 0 & 0 & 0 & 0 & 0 & 0 & 0 & 0 \\
SlotVLA (h=8) & LT & 0 & 0 & 0 & 0 & 0 & 0 & 0 & 0 & 0 & 0 & 0 \\
Naive E-SlotSSM (oracle) & LT & 0 & 0 & 0 & 0 & 0 & 0 & 0 & 0 & 0 & 0 & 0 \\
\bottomrule
\end{tabular}
\end{table*}
\begin{table*}[!htbp]
\centering\scriptsize
\setlength{\tabcolsep}{3pt}
\caption{LIBERO-Mem subgoal completion (rows 1--8).}
\begin{tabular}{@{}p{4.1cm}lrrrrrrrrrrr@{}}
\toprule
Method & Source & T1 & T2 & T3 & T4 & T5 & T6 & T7 & T8 & T9 & T10 & Avg. \\
\midrule
SmoLSTM (flow) & SL & -- & -- & -- & -- & -- & -- & -- & -- & -- & -- & 85.1 \\
\textbf{AgentMem (current)} & \textbf{AgentMem} & \textbf{75} & \textbf{86} & \textbf{60.33} & \textbf{81.67} & \textbf{68} & \textbf{56.43} & \textbf{82} & \textbf{78.75} & \textbf{83} & \textbf{99.5} & \textbf{77.07} \\
TRP & TR & -- & -- & -- & -- & -- & -- & -- & -- & -- & -- & 59.7 \\
TRP-perturbed & TR & -- & -- & -- & -- & -- & -- & -- & -- & -- & -- & 48 \\
TRP-no-carry & TR & -- & -- & -- & -- & -- & -- & -- & -- & -- & -- & 21.6 \\
SlotSSM [oracle SG] & LM & 50 & 0 & 33.3 & 0 & 14.3 & 0 & 0 & 0 & 30 & 20 & 14.8 \\
SlotVLA (8 frames) & LM & 50 & 0 & 0 & 0 & 0 & 0 & 0 & 0 & 0 & 0 & 5 \\
pi0 & LM & 50 & 0 & 0 & 0 & 0 & 0 & 0 & 0 & 0 & 0 & 5 \\
\bottomrule
\end{tabular}
\end{table*}
\begin{table*}[!htbp]
\centering\scriptsize
\setlength{\tabcolsep}{3pt}
\caption{RMBench success rates (rows 1--21).}
\begin{tabular}{@{}p{4.1cm}lrrrrrrrrrrrr@{}}
\toprule
Method & Source & Obs & Rearr & Back & SwapB & SwapT & Batt & Rank & Cover & Press & Mat & 9-task & 10-task \\
\midrule
SimpleMemVLA & SM & 65 & 100 & 100 & 100 & 93 & 90 & 100 & 98 & 100 & 100 & 94 & 94.6 \\
MemoryWAM & SM & 27 & 100 & 100 & 100 & 94 & 41 & 100 & 98 & 87 & -- & 83 & -- \\
EvoMem-VLA & EM & 28 & 100 & 92 & 100 & 100 & 44 & 96 & 96 & 70 & -- & 80.7 & -- \\
DIM-WAM & SM & 13 & 99 & 98 & 96 & 97 & 48 & 87 & 56 & 34 & -- & 69.8 & -- \\
EventVLA (VA) & EV,SM,EM & 21 & 96 & 95 & 96 & 87 & 35 & 81 & 97 & 3 & -- & 67.8 & -- \\
\textbf{AgentMem (current)} & \textbf{AgentMem} & \textbf{47} & \textbf{60} & \textbf{98} & \textbf{59} & \textbf{76} & \textbf{35} & \textbf{91} & \textbf{65} & \textbf{30} & \textbf{66} & \textbf{62.33} & \textbf{62.7} \\
Mem4Act & MA & 22 & 99 & 100 & 100 & 63 & 29 & 26 & 100 & 10 & -- & 61 & -- \\
NativeMEM & MA & 44 & 89 & 100 & 94 & 52 & 14 & 11 & 32 & 2 & -- & 48.7 & -- \\
Mem-0 & EV,SM,EM,MA & 4 & 89 & 90 & 67 & 14 & 28 & 18 & 68 & 0 & -- & 42 & -- \\
MemoryVLA (QwenOFT) & EV,EM & 2 & 53 & 81 & 76 & 9 & 33 & 53 & 69 & 0 & -- & 41.7 & -- \\
EventVLA (w/o initial) & EV & 10 & 64 & 63 & 16 & 8 & 39 & 87 & 15 & 2 & -- & 33.7 & -- \\
HiMem-WAM & SM & 28 & 33 & 32 & 38 & 27 & 28 & 24 & 19 & 8 & -- & 26.3 & -- \\
EventVLA (w/o short-term) & EV & 15 & 34 & 20 & 18 & 94 & 16 & 14 & 4 & 0 & -- & 23.8 & -- \\
HAMLET & MA & 12 & 52 & 44 & 22 & 31 & 15 & 5 & 30 & 0 & -- & 23.4 & -- \\
MemoryVLA (OpenVLA) & EV,EM & 0 & 22 & 50 & 17 & 9 & 25 & 12 & 40 & 0 & -- & 19.4 & -- \\
pi0.5 & EV,SM,EM,MA & 9 & 13 & 11 & 24 & 15 & 16 & 6 & 0 & 0 & -- & 10.4 & -- \\
X-VLA & EV,SM,EM,MA & 9 & 13 & 18 & 16 & 3 & 26 & 1 & 2 & 0 & -- & 9.8 & -- \\
MemER & EV,EM & 7 & 17 & 0 & 14 & 7 & 27 & 0 & 6 & 0 & -- & 8.7 & -- \\
ACT & EV,SM,EM,MA & 1 & 29 & 0 & 2 & 2 & 19 & 0 & 0 & 0 & -- & 5.9 & -- \\
DP & EV,SM,EM,MA & 1 & 0 & 0 & 11 & 20 & 10 & 10 & 0 & 0 & -- & 5.8 & -- \\
QwenOFT & EV,EM & 0 & 0 & 0 & 0 & 0 & 14 & 37 & 0 & 0 & -- & 5.6 & -- \\
\bottomrule
\end{tabular}
\end{table*}
\begin{table*}[!htbp]
\centering\small
\setlength{\tabcolsep}{3pt}
\caption{Selected RoboDojo Memory references (SR, \%). Cover: cover blocks; SwapB: swap blocks; SwapT: swap T; Press: press by number; Imit: imitate sorting sequence; Conv: match and pick from conveyor. We retain representative baselines and higher-scoring references rather than the full leaderboard.}
\label{tab:robodojo-selected}
\begin{tabular}{@{}p{4.1cm}lrrrrrrr@{}}
\toprule
Method & Source & Cover & SwapB & SwapT & Press & Imit & Conv & Avg. \\
\midrule
HexaShell (no skills) & HX & 100 & 100 & 100 & 96.67 & 96.67 & 100 & 98.89 \\
Q0-Preview & DJ & 100 & 90 & 100 & 92 & 96 & 84 & 93.67 \\
Agent as Policy (seed 0) & AP & 96 & 96 & 94 & 84 & 80 & 96 & 91 \\
GPT-6 Astra & DJ,UP & 46 & 0 & 18 & 70 & 36 & 62 & 38.67 \\
SimpleMemVLA & DJ,SM & 91 & 11 & 1 & 57 & 0 & 39 & 33.17 \\
\textbf{AgentMem (current)} & \textbf{AgentMem} & \textbf{70} & \textbf{6} & \textbf{4} & \textbf{48} & \textbf{0} & \textbf{54} & \textbf{30.33} \\
Hy-Embodied-0.5 & DJ & 27.33 & 2.67 & 0 & 22.67 & 0 & 20 & 12.11 \\
EventVLA & DJ & 0 & 14.67 & 0 & 2.67 & 0 & 11.33 & 4.78 \\
pi0.5 & DJ & 13.33 & 0 & 0.67 & 0 & 0 & 14 & 4.67 \\
SmolVLA & DJ & 4.67 & 0 & 0 & 0 & 0 & 10 & 2.45 \\
pi0 & DJ & 6 & 0 & 0 & 0.67 & 0 & 6 & 2.11 \\
DeepSeek-Flash & DJ,UP & 0 & 0 & 0 & 0 & 0 & 10 & 1.67 \\
GPT-5.5 & DJ,UP & 0 & 0 & 0 & 0 & 0 & 10 & 1.67 \\
ACT & DJ & 0.67 & 0 & 0 & 0 & 0 & 4.67 & 0.89 \\
OpenVLA-OFT & DJ & 0 & 0 & 0 & 0 & 0 & 0 & 0 \\
\bottomrule
\end{tabular}
\end{table*}
\clearpage
\endgroup

\end{document}